\documentclass[10pt,journal,compsoc]{IEEEtran}
\usepackage[cmex10]{amsmath}
\usepackage{amsfonts}
\usepackage{algorithmic}
\usepackage{algorithm}
\usepackage{graphicx}
\usepackage{array}
\usepackage{booktabs}
\usepackage{epstopdf}
\usepackage{arydshln}

\usepackage{url}
\usepackage{color}
\usepackage{mathrsfs} 
\usepackage[switch]{lineno}

\usepackage[pagebackref=true,breaklinks=true,citecolor=blue,linkcolor=blue,letterpaper=true,colorlinks,bookmarks=false]{hyperref}

\begin{document}

\title{Visual Tracking via Dynamic Graph Learning}

\author{Chenglong~Li,
        Liang~Lin,
        Wangmeng~Zuo,
		  Jin~Tang,
        and Ming-Hsuan~Yang
\thanks{C. Li and J. Tang are with the School of Computer Science and Technology, Anhui University, Hefei, 230601, China. C. Li is also with the Center for Research on Intelligent Perception and Computing (CRIPAC), National Laboratory of Pattern Recognition (NLPR), Institute of Automation, Chinese Academy of Sciences (CASIA), Beijing, 100190, China. (e-mail: lcl1314@foxmail.com; jtang99029@foxmail.com)}
\thanks{L. Lin is with the School of Data and Computer Science, Sun Yat-Sen University, Guangzhou, 510006, China. (e-mail: linliang@ieee.org)}
\thanks{W. Zuo is with the School of Computer Science and Technology, Harbin Institute of Technology, Harbin, 150001, China. (e-mail: cswmzuo@gmail.com)}
\thanks{M.-H. Yang is with the School of Engineering, University of California, Merced, CA 95344 USA. (e-mail: mhyang@ucmerced.edu)}
}

\markboth{IEEE Transactions on Pattern Analysis and Machine Intelligence}
{}

\IEEEtitleabstractindextext{%
\begin{abstract}
Existing visual tracking methods usually localize a target object with a bounding box, in which the performance of the foreground object trackers or detectors is often 
affected by the inclusion of background clutter. 
To handle this problem, we learn a patch-based graph representation for visual tracking.
The tracked object is modeled by with a graph by taking a set of non-overlapping image patches as nodes, 
in which the weight of each node indicates how likely it belongs to the foreground and edges are weighted for indicating 
the appearance compatibility of two neighboring nodes. 
This graph is dynamically learned 
and applied in object tracking and model updating. 
During the tracking process, the proposed algorithm performs three main steps in each frame. 
%
First, the graph is initialized by assigning binary weights of some image patches to indicate the object and background patches according to the predicted bounding box. 
Second, the graph is optimized to refine the patch weights by using a novel 
alternating direction method of multipliers.
Third, the object feature representation is updated by imposing the weights of patches on the extracted image features. 
The object location is predicted by maximizing the classification score in the structured
support vector machine. 
Extensive experiments show that the proposed tracking algorithm performs well against 
the state-of-the-art methods on large-scale benchmark datasets. 
\end{abstract}

}

\maketitle

\IEEEdisplaynontitleabstractindextext

\IEEEpeerreviewmaketitle

\section{Introduction}\label{sec::introduction}

\IEEEPARstart{V}{isual} tracking is a fundamental and active research topic in computer vision due to its wide range of applications such as activity analysis, visual surveillance and self-driving systems. 
Despite significant progress has been made in recent years, it remains a challenging issue, 
partly due to the difficulty of constructing robust object representation to cope with various 
factors including camera motion, partial occlusion, background clutter and illumination change.

Numerous visual tracking methods recently adopt the tracking-by-detection paradigm, i.e., separating the foreground object from the background over time using a classifier. 
These methods usually localize the object using a bounding box, and draw positive (negative) samples from inside (outside) of the bounding box for the classifier update. 
Since the ground-truth object labeling is only available in the initial frame, incrementally updating the object classifier in subsequent frames often result in model drift 
due to inclusion of outlier samples.

Significant efforts have been made to alleviate the effects of outlier samples in visual tracking~\cite{Comaniciu03pami,Fragtrack06cvpr,Stuck11iccv,He13cvpr,MEEM14eccv,Zhang14cvpr,Kim15iccv,RPT15cvpr,Timofte16cviu}. 
Several methods in~\cite{Comaniciu03pami,Stuck11iccv,He13cvpr} update the object classifiers by considering the distances of samples with respect to the bounding box center, e.g., the samples close to the center receiving higher weights. 
Some other methods~\cite{Duffner13iccv,Yang14tip} segment foreground objects from the background during the tracking process to exclude background clutter. 
However, these methods are limited in dealing with cluttered backgrounds (e.g., unreliable segmented object masks). 
To improve the robustness, Kim et al.~\cite{Kim15iccv} define an image patch based 8-neighbor graph to represent the tracked object, in which if two nodes are connected by an edge if they are are $8$-neighbors and the edge weight is computed based on  low-level feature distance. 
There are two main issues with this approach: i) it only considers the spatial neighbors and do not capture the intrinsic relationship between patches; ii) it uses low-level feature which are less effective in the presence of clutter and noise. 

To handle these issues, we learn a robust object representation for visual tracking. 
Given one bounding box of the target object, we partition it into non-overlapping local patches, which are described by color and gradient histograms. 
Instead of using static structures in existing methods~\cite{Yang13cvpr,Kim15iccv}, we learn a dynamic graph with patches as nodes (i.e., adaptive structure and node weights for each frame) for representing the target object, where the weight of each node describes how likely it belongs to the target object, and the edge weight indicates the appearance compatibility of two neighboring patches. 
Existing methods usually perform two steps for node weight computation, i.e.,  
first constructing the graph with a static structure and low-level features, and then computing 
node weights based on some semi-supervised methods~\cite{Yang13cvpr,Kim15iccv}. 
In this work, we propose a novel representation model to jointly 
learn the graph that infers the graph structure, edge weights and node weights.

With the advances of compressed sensing~\cite{cs06tit}, numerous methods exploiting the relationship of data representations have been proposed~\cite{Elhamifar09cvpr,Liu13pami,Zhuang12cvpr,Guo15ijcai}. 
The representations are generally utilized to define~\cite{Elhamifar09cvpr,Liu13pami,Zhuang12cvpr} or learn~\cite{Guo15ijcai} the affinity matrix of a graph. 
Motivated by these methods, we represent each patch descriptor as a linear combination of other patch descriptors, and develop a model to jointly optimize the graph structure, edge weights and node weights while suppressing the effects of noise from clutter and low-level features.

It is worth mentioning that our model has the following three distinctive properties: 
1) it is capable of collaboratively optimize the graph structure, edge weights and node weights according to the underlying intrinsic relationship, which provides a flexible solution for visual tracking and other vision problems such as saliency detection~\cite{Yang13cvpr} and semi-supervised object segmentation~\cite{Li16tip}; 
2) it is effective to suppress the effects of noise from pixels and low-level features in computing the affinity matrix of the graph; 
3) it is generic, and can incorporate other constraints (e.g., low-rank and sparse constraints) to further improve the robustness of graph learning.

To improve the tracking efficiency, we develop an alternating direction method of multipliers (ADMM) algorithm to seek the solution of the proposed model. 
In particular, the alternating direction method~\cite{Lin11nips} 
is used to linearize the quadratic penalty term while avoiding an auxiliary variable and some matrix inversions such that  each subproblem can be efficiently solved with a 
closed-form solution. 
We construct the robust object representations by combining patch features with the optimized weights, and then apply the structured support vector machine (SVM)~\cite{Stuck11iccv} for object tracking and model update.

In each frame, the proposed algorithm is carried out with several steps. 
First, the graph is initialized with binary weights to according to the ground truth 
(first frame) or the predicted bounding box (subsequent frames). 
Second, the graph is optimized by a linearized ADMM algorithm. 
Third, the object feature representation is updated by imposing the patch weights 
on the extracted image features. 
The object location is finally predicted by adopting the structured SVM.

We make three major contributions for visual tracking and related applications in this work:
\begin{itemize}
\item We propose an effective approach to alleviate the effects of background clutter 
in visual tracking. 
Extensive experiments show that the proposed method outperforms most 
state-of-the-art trackers on four benchmark datasets.

\item We present a novel representation model to learn a dynamic graph according to the intrinsic relationship among image patches. 
The proposed model is jointly optimizes the graph structure, edge weights and node weights while suppressing the effects of patch noise and/or corruption. 
It also provides a general solution for visual tracking and other vision problems such as saliency detection~\cite{Yang13cvpr,PISA15tip} and interactive object segmentation~\cite{Li16tip,Li15cvpr}.

\item We develop an ADMM algorithm to efficiently solve the associated optimization problem. 
Empirically, the proposed optimization algorithm  exhibits stable convergence behavior on  real image data. 

\end{itemize}

This paper provides a more complete understanding of the early results~\cite{Li17aaai}, with more background, insights, analysis, and evaluation. 
In particular, our approach advances the early work in several aspects. 
First, we utilize the data representations to learn more meaningful graph affinity, 
instead of directly using data representations. 
Second, we generalize the graph learning algorithm to incorporating different constraints (or priors), such as the low rank, sparse and spatial smoothness constraints. 
We further discuss the merits for graph learning, and instantiate the sparse constraints into our framework. 
Third, scale estimation is considered in this work to improve visual tracking 
Finally, we carry out extensive experiments on large-scale benchmark datasets 
to demonstrate the effectiveness of the proposed algorithm, including quantitative comparisons with the state-of-the-art trackers and ablation studies. 

\section{Literature Review}
\label{sec::related_work}
A plethora of visual methods have been proposed in the literature~\cite{RGBbenchmark15pami, NUS-PRO16pami} and  we  discuss  the most related work in this section.
We discuss the advances of visual tracking in two aspects: 
constructing robust appearance models to alleviate the effect of background clutter, 
and learning data affinity for model construction.

\subsection{Appearance Models}
Various tracking methods have been proposed to improve the robustness to nuisance factors including label ambiguity, background clutter, corruption and occlusion. 
Grabner et al.~\cite{Grabner08eccv} present a tracking approach 
that adapts to drastic appearance changes and limits the drifting problem.
%
The knowledge from labeled data is used to construct static prior for online classifier while unlabeled samples are explored in a principled manner during tracking. 
Babenko et al.~\cite{MIT11pami} use a bag of multiple samples, instead of a single sample, 
to update the classifier reliably. 
To avoid the label ambiguity, Hare et al.~\cite{Stuck11iccv} exploit
structured samples instead of binary-labeled samples when training the classifier 
in the structured SVM framework~\cite{Tsochantaridis05jmlr}.

To alleviate the effects of background clutter, one representative approach 
is to assign weights to different pixels or patches in the bounding box. 
Comaniciu et al.~\cite{Comaniciu03pami} develop the kernel-based method to assign smaller weights to boundary pixels for histogram matching. 
In~\cite{He13cvpr} He et al. also assume that pixels far from a box center should be less important. 
These methods do not perform well when a target shape cannot be well described by a rectangle or occluded. 
Some methods~\cite{Duffner13iccv,Yang14tip} integrate segmentation results in visual tracking to alleviate the effects of background. 
These algorithms, however, reply heavily on the quality of segmentation results. 
Kim et al.~\cite{Kim15iccv} develop a random walk restart algorithm on a 8-neighbor graph to compute patch weights within the target object bounding box. 
Nevertheless, the constructed graph does not  capture the relationship between patches well.

\subsection{Data Affinity}
In vision and learning problems, we often have a set of data ${\bf X}=\{{\bf x}_1, \ldots,{\bf x}_n\}\in\mathbb{R}^{d\times n}$ drawn from a union of $c$ subspaces $\{\mathbb{S}_{s=1}^c\}$, where $d$ is the feature dimension and $n$ is the number of data vectors. 
To characterize the relation between the data in ${\bf X}$, the key is to construct an effective affinity matrix ${\bf A}\in\mathbb{R}^{n\times n}$, in which ${\bf A}_{ij}$ reflects the similarity between data points ${\bf X}_i$ and ${\bf X}_j$. 
While computing Euclidean distances on the raw data is the most intuitive way to construct the data affinity matrix, such metric usually does not reveal the global subspace structure of data well.

With the advances of compressed sensing~\cite{cs06tit}, significant efforts 
have been made to exploit the relationship of data representations~\cite{Elhamifar09cvpr,Liu13pami,Zhuang12cvpr,Guo15ijcai}
where the general formulation is described by:
\begin{equation}
\label{eq:general_model}
\begin{aligned}
&\underset{{\bf Z}, {\bf E}}{\min}~
\alpha\Theta({\bf Z})+\beta\Phi({\bf E})\\
&~\mbox{s.t.} \quad {\bf X}={\bf XZ} + {\bf E},
\end{aligned}
\end{equation}
where ${\bf Z}\in\mathbb{R}^{n\times n}$ and ${\bf E}\in\mathbb{R}^{d\times n}$ denote the representation matrix and the residual matrix, respectively. 
In~\eqref{eq:general_model}, $\Theta({\bf Z})$ is the regularizer on ${\bf Z}$. $\Phi({\bf E})$ is
 the model of ${\bf E}$, which can be with different forms depending on data characteristics; and
 $\alpha$ as well as $\beta$ are the weight parameters.

\begin{figure*}[t]
\centering
\includegraphics[width = 2\columnwidth]{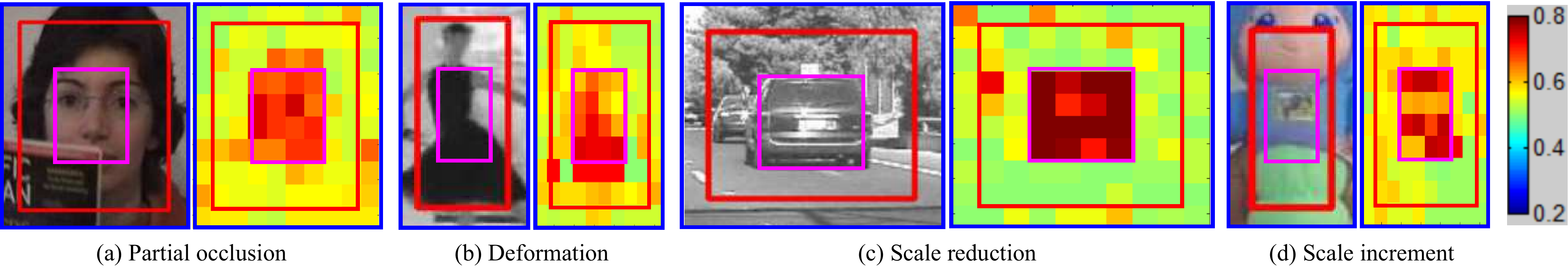}
\caption{Illustration of the original, shrunk and expanded bounding boxes on 4 video frames with different challenges, which are represented by the red, pink and blue colors, respectively. The optimized patch weights are also shown for clarity, in which the hotter color indicates the larger weight. One can see that the optimized patch weights are beneficial to suppressing the effects of background clutter.  }\label{fig::weight_sample}
\end{figure*}

Numerous methods have been developed to extract compact information from image data including sparse representation (SR)~\cite{Elhamifar09cvpr} 
\begin{equation}
\Theta({\bf Z})=||{\bf Z}||_0,
\end{equation}
and low-rank representation (LRR)~\cite{Liu13pami},
\begin{equation}
\Theta({\bf Z})=\mbox{rank}({\bf Z}).
\end{equation}
%
Different from traditional methods, SR schemes can be used to exploit higher order relationships among more data points effectively, and hence provide more compact and discriminative
models~\cite{Zhuang12cvpr}. 
The main drawback of SR methods is that data is processed individually without taking the existence of inherent global structure into account. 
On the other hand, low-rank representation models use low rank constraints on data representations to capture the global structure of the whole data. 
It has been shown that, under mild conditions, LRR methods can preserve the membership of samples that belong to the same subspace well. 
Recently, Zhuang et al.~\cite{Zhuang12cvpr} harnesses both sparsity and low-rankness of data to learn more informative representations.

In general, after solving the problem~\eqref{eq:general_model}, the representation is used to define the affinity matrix of an undirected graph with ${\bf a}_{ij} = \frac{{\bf z}_{ij}+{\bf z}_{ji}}{2}$ for ${\bf x}_i$ and ${\bf x}_j$.
%
However, the metric implies the affinity is already not the same as the original definition. This is because the affinity defines an approximation to the pairwise distances between data samples while the representation is the reconstruction coefficients of one sample from others. 
As such, Guo~\cite{Guo15ijcai} proposes a method to simultaneously learn data representations and affinity matrix.
Experimental results on the synthetic and real datasets demonstrate the effectiveness of learning model representation and affinity matrix jointly. 

\section{Patch-based Graph Learning}
\label{sec::model}
Given one bounding box that encloses the target object, we partition it into non-overlapping patches and assign each one with a weight that reflects the importance in describing the target object to alleviate the effects of background clutter. 
We concatenate these weighted patch descriptors into a feature vector and use the Struck~\cite{Stuck11iccv} method for object tracking. 
In this section, we first describe a sparse low-rank model based on local patches, and then an efficient ADMM algorithm to compute the weights. 

\subsection{Formulation}
Each bounding box of the target object is partitioned into $n$ non-overlapping patches, and a set of low-level appearance features are extracted and combined into one single $d$-dimensional feature vector ${\bf x}_i$ for characterizing the $i$-th patch. 
Using these patches as graph nodes, each bounding box can be represented with a graph, in which the weight of each node describes how likely it belongs to the target object
and the edge weight between two neighboring patches indicates appearance compatibility.

For visual tracking, some patches in a target bounding box may belong to background due to irregular shape, scale variation and partial occlusion of the target object, 
as shown in Figure~\ref{fig::weight_sample}. 
Thus, we assign a weight for each graph node to alleviate the effects of background pixels on object tracking and model update. 
On the other hand, instead of constructing spatially ordered graphs~\cite{Yang13cvpr,Kim15iccv}, 
the edges are dynamically learned for capturing the intrinsic relationship of data.
In this work, we propose a novel graph learning approach to infer the edges and  node weights jointly which performs well against the state-of-the-art alternatives for visual tracking.

All the feature vectors of $n$ patches in one bounding box form the data matrix ${\bf X}=[{\bf x}_1,{\bf x}_2, \ldots,{\bf x}_n]\in\mathbb{R}^{d\times n}$. 
Each patch descriptor can be represented as a linear combination of remaining patch descriptors, and the representation of all patch vectors can then be formulated by ${\bf X}={\bf XZ}$, where ${\bf Z} \in \mathbb{R}^{n\times n}$ is the representation coefficient matrix. Since the patch feature matrix often contains noise, the representation can be obtained by solving the objective function~\eqref{eq:general_model}.

The optimal representation coefficient matrix in~\eqref{eq:general_model} is 
often utilized to define the affinity matrix of an undirected graph in the way of $\frac{|{\bf z}_{ij}|+|{\bf z}_{ji}|}{2}$ for the feature vector ${\bf x}_i$ and ${\bf x}_j$. 
%
As ${\bf z}_{ij}$ and ${\bf z}_{ji}$ are the reconstruction coefficients, this encoded information is not the same as the original definition, which defines an approximation to the pairwise distances between ${\bf x}_i$ and ${\bf x}_j$~\cite{Guo15ijcai}.
Therefore, we also learn the affinity matrix by assuming that the patch features should have  larger probabilities to be in the same cluster if their representations have smaller distance, and impose the following constraints,
\begin{equation}
\min\limits_{{\bf A1}={\bf 1},{\bf A}\geq 0}\sum_{i,j=1}^n||{\bf z}_i-{\bf z}_j||_F^2{\bf a}_{ij},
\end{equation} 
where ${\bf A}\in\mathbb{R}^{n\times n}$ is the desired affinity matrix,
whose element ${\bf a}_{ij}$ reflects the probability of the patch features ${\bf x}_i$ and ${\bf x}_j$ from the same cluster based on the distance between their representations ${\bf z}_i$ and ${\bf z}_j$. 
The constraints ${\bf A1}={\bf 1}$ and ${\bf A}\geq 0$ guarantee the probability property of each column of ${\bf A}$. 
With some simple algebra, we integrate these constraints into~\eqref{eq:general_model}, and have
\begin{equation}
\label{eq:general_model1}
\begin{aligned}
&\underset{{\bf Z}, {\bf E},{\bf A}}{\min}~
\alpha\Theta({\bf Z})+\beta\Phi({\bf E})+\gamma~\mbox{tr}({\bf Z}{\bf L}_{\bf A}{\bf Z}^{\top})+\frac{\lambda}{2}||{\bf A}||_F^2\\
&~~\mbox{s.t.}  \quad {\bf X}={\bf XZ} + {\bf E},~{\bf A1}={\bf 1},~{\bf A}\geq 0,
\end{aligned}
\end{equation}
where ${\bf L}_{\bf A}={\bf D}_{\bf A}-{\bf A}$ is the Laplacian matrix of ${\bf A}$, and ${\bf D}_{\bf A}$ is the degree matrix of ${\bf A}$, a diagonal matrix whose the $i$-th diagonal element is $\sum_j{\bf a}_{ij}$. 
In~\eqref{eq:general_model1},
$\gamma$ and $\lambda$ are weight parameters. 
In addition, the last term is used to avoid overfitting.
{Note that minimizing the term $\mbox{tr}({\bf Z}{\bf L}_{\bf A}{\bf Z}^{\top})$ could exclude the trivial solution ${\bf Z}={\bf I}$, where ${\bf I}$ indicates the identity matrix.
The trivial solution ${\bf E}={\bf 0}$ is also not achieved as it means the data are clean, which is an ``ideal'' case, and does not exist in real-world applications.}

\begin{figure}[t]
\centering
\includegraphics[width = \columnwidth]{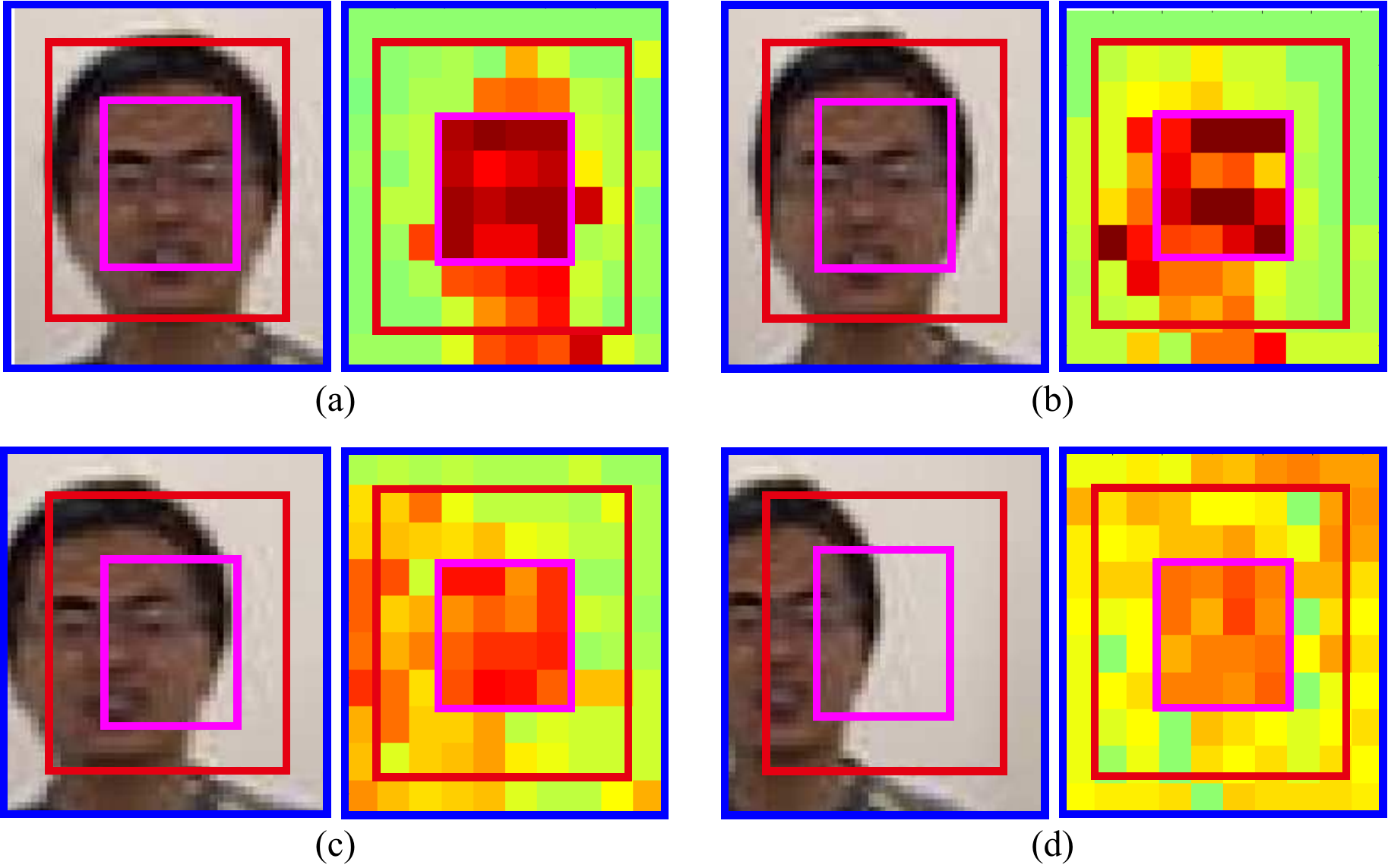}
\caption{Illustration of clutter and noise effects of initial seeds on a video frame in the sequence $boy$~\cite{RGBbenchmark15pami}, please refer to Figure~\ref{fig::weight_sample} for the detailed descriptions. One can see that the optimized patch weights are robust to clutter and noise, but bad if most of initial seeds are clutter and noise.  }\label{fig::weight_sample1}
\end{figure}

To alleviate the effects of background clutter, we assign a weight ${\bf w}_i$ for each patch $i$ using a semi-supervised formulation. 
Let ${\bf r}=\{{\bf r}_1,{\bf r}_2, \ldots,{\bf r}_n\}^{\top}$ be an initial weight vector, in which ${\bf r}_i=1$ if ${\bf r}_i$ is a target object patch, and ${\bf r}_i=0$ indicating a background patch. 
In this work, ${\bf r}$ is computed by the initial ground truth (for first frame) or the previous tracking result (for subsequent frames).
For $i$-th patch, if it belongs to the shrunk region of the bounding box then ${\bf r}_i$ is $1$, and if it belongs to the expanded region of the bounding box then ${\bf r}_i$ is $0$. 
Figure~\ref{fig::weight_sample} shows the one example how the weights are assigned. 
Although using a simple initialization strategy, we demonstrate empirically this scheme performs well empirically, and show the robustness to clutter and noise in Figure~\ref{fig::weight_sample1}. 

The remaining patches are non-determined, and are diffused by other patches. 
To this end, we define an indicator vector $\Gamma$ that $\Gamma_i=1$ indicates the $i$-th patch is foreground or background patch, and $\Gamma_i=0$ denotes the $i$-th patch is non-determined patch. We integrate the patch weights into~\eqref{eq:general_model1}, and obtain
\begin{equation}
\label{eq:graph_model1}
\begin{aligned}
&\underset{{\bf Z}, {\bf E}, {\bf A}, {\bf w}}{\min}~\alpha\Theta({\bf Z})+\beta\Phi({\bf E})+\gamma~\mbox{tr}({\bf Z}{\bf L}_{\bf A}{\bf Z}^{\top})\\
&+\lambda_1\sum_{i,j}{{\bf a}_{ij}({\bf w}_i-{\bf w}_j)^2}+\frac{\lambda_2}{2}||\Gamma\circ({\bf w}-{\bf r})||^2\\
&+\frac{\lambda_3}{2}||{\bf A}||_F^2+\frac{\lambda_4}{2}||{\bf w}||^2\\
&\mbox{s.t.} \quad {\bf X}={\bf XZ} + {\bf E},~{\bf A1}={\bf 1},~{\bf A}\geq 0,~{\bf w}\geq 0,
\end{aligned}
\end{equation}
where $\circ$ indicates the element-wise product. $\lambda_1$, $\lambda_2$, $\lambda_3$ and $\lambda_4$ are weight parameters. The third and fourth terms are the smoothness 
and fitting constraints. 
Since the indicator vector $\Gamma$ removes fitness constraint of non-determined patch weights, we introduce the last term to avoid overfitting. 
{Specifically, the smoothness term of ${\bf w}$ constrains that ${\bf w}_i$ and ${\bf w}_j$ are similar to each other when ${\bf a}_{ij}$ is non-zero, and the fitting term of ${\bf w}$ controls that its elements are close to 0 or 1. However, the fitting constraint is partial, and we thus introduce $||{\bf w}||^2$ to avoid its element amplitude too large.}

\subsection{Discussion}
As discussed in Section~\ref{sec::related_work}, the regularizer $\Theta({\bf Z})$ is 
usually based on sparse or low-rank priors, e.g., sparse representation (SR) and low-rank representation (LRR). 
The SR methods exploit higher order relationships among more data points 
and hence is more discriminative~\cite{Elhamifar09cvpr,Wright10procIEEE}. 
The LRR approaches employ low rank constraints on data representations to capture the global structure 
of data points, and thus is robust to noise and corruption~\cite{Zhuang12cvpr,Liu13pami}. 
However, the LRR methods require singular value decomposition (SVD) operations at each iteration, which is computationally demanding. 
%
%
Therefore, we impose the sparse constraints (i.e., $\ell_1$-norm, a convex surrogate for $\ell_0$-norm) on ${\bf Z}$ in this work for computational efficiency.

In \eqref{eq:graph_model1}, 
the model $\Phi({\bf E})$ can be in different forms based on the characteristic of data. 
For visual tracking, as some image patches are corrupted (e.g., occluded by background or other objects), 
we employ $\ell_{2,1}$-norm (a convex surrogate for $\ell_{2,0}$-norm) on ${\bf E}$. 
Putting the data terms and prior together, we have:
\begin{equation}
\label{eq:graph_model2}
\begin{aligned}
&\underset{{\bf Z}, {\bf E}, {\bf A}, {\bf w}}{\min}~\alpha||{\bf Z}||_1+\beta||{\bf E}||_{2,1}+\gamma~\mbox{tr}({\bf Z}{\bf L}_{\bf A}{\bf Z}^{\top})\\
&+\lambda_1\sum_{i,j}{{\bf a}_{ij}({\bf w}_i-{\bf w}_j)^2}+\frac{\lambda_2}{2}||\Gamma\circ({\bf w}-{\bf r})||^2\\
&+\frac{\lambda_3}{2}||{\bf A}||_F^2+\frac{\lambda_4}{2}||{\bf w}||^2\\
&\mbox{s.t.} \quad {\bf X}={\bf XZ} + {\bf E},~{\bf A1}={\bf 1},~{\bf A}\geq 0,~{\bf w}\geq 0,
\end{aligned}
\end{equation}
where $||\cdot||_1$ and $||\cdot||_{2,1}$ denote $\ell_1$-norm and $\ell_{2,1}$-norm of a matrix, respectively.{~\eqref{eq:graph_model2} is reasonable as two patches should prefer to be sparsely represented by same set of patches if they are similar. In particular, for optimizing ${\bf Z}$, we exploit higher order relationship among patches by minimizing $l_1$ norm on ${\bf Z}$, and also penalty inconsistency between ${\bf z}_i$ and ${\bf z}_j$ when patch features ${\bf x}_i$ and ${\bf x}_j$ are similar (i.e., large ${\bf a}_{ij}$) by minimizing $\mbox{tr}({\bf Z}{\bf L}_{\bf A}{\bf Z}^{\top})$.}

{It is worth noting that although ${\bf A}$ is a non-symmetrical affinity matrix, as shown in next section, the solutions of the variables that rely on ${\bf A}$ (i.e., ${\bf Q}$ and ${\bf w}$) are based on a symmetrical affinity matrix, i.e., $({\bf A}+{\bf A}^{\top})/2$.}

\subsection{Optimization}
%
Although~\eqref{eq:graph_model2} is not jointly convex on ${\bf Z}$, ${\bf E}$, ${\bf A}$ and ${\bf w}$, but it is convex with respect to each of them when others are fixed. 
The ADMM (Alternating Direction Method of Multipliers) algorithm~\cite{Lin11nips} has shown to be an efficient and effective solver of problems similar to~\eqref{eq:graph_model2}. 
To apply ADMM for the above problem, we need to make the objective function separable. 
Therefore, we introduce an auxiliary variable ${\bf Q}\in\mathbb{R}^{n\times n}$ to replace ${\bf Z}$ in~\eqref{eq:graph_model2}:
\begin{equation}
\label{eq:graph_model1_separable}
\begin{aligned}
&\underset{{\bf Z}, {\bf E}, {\bf A}, {\bf w}, {\bf Q}}{\min}~\alpha||{\bf Z}||_1+\beta||{\bf E}||_{2,1}+\gamma~\mbox{tr}({\bf Q}{\bf L}_{\bf A}{\bf Q}^{\top})\\
&+\lambda_1\sum_{i,j}{{\bf a}_{ij}({\bf w}_i-{\bf w}_j)^2}+\frac{\lambda_2}{2}||\Gamma\circ({\bf w}-{\bf r})||^2\\
&+\frac{\lambda_3}{2}||{\bf A}||_F^2+\frac{\lambda_4}{2}||{\bf w}||^2\\
&\mbox{s.t.} \quad {\bf X}={\bf XZ} + {\bf E},~{\bf Z}={\bf Q},{\bf A1}={\bf~ 1},{\bf A}\geq 0,~{\bf w}\geq 0.
\end{aligned}
\end{equation}

The augmented Lagrangian function of~\eqref{eq:graph_model1_separable} is
\begin{equation}
\label{eq:graph_model1_ALM}
\begin{aligned}
&\mathbb{L}_{\{{\bf A1}={\bf 1},{\bf A}\geq 0,{\bf w}\geq 0\}}({\bf Z},{\bf Q},{\bf E},{\bf A},{\bf w})\\
&=\alpha||{\bf Z}||_1+\beta||{\bf E}||_{2,1}+\gamma~\mbox{tr}({\bf Q}{\bf L}_{\bf A}{\bf Q}^{\top})\\
&+\lambda_1\sum_{i,j}{{\bf a}_{ij}({\bf w}_i-{\bf w}_j)^2}+\frac{\lambda_2}{2}||\Gamma\circ({\bf w}-{\bf r})||^2\\
&+\frac{\lambda_3}{2}||{\bf A}||_F^2+\frac{\lambda_4}{2}||{\bf w}||^2-\frac{1}{2\mu}(||{\bf Y}_1||_F^2+||{\bf Y}_2||_F^2)\\
&+f({\bf Z},{\bf Q},{\bf E},{\bf w},{\bf Y}_1,{\bf Y}_2,\mu),
\end{aligned}
\end{equation}
where $\mu>0$ is the penalty parameter, and $f({\bf Z},{\bf Q},{\bf E},{\bf w},{\bf Y}_1,{\bf Y}_2,\mu)=\frac{\mu}{2}(||{\bf X}-{\bf XZ}-{\bf E}+{\bf Y}_1/\mu||_F^2+||{\bf Z}-{\bf Q}+{\bf Y}_2/\mu||_F^2)$.
In the above equation,  ${\bf Y}_1$ and ${\bf Y}_2$ are the Lagrangian multipliers. The ADMM alternatively updates one variable by minimizing $\mathbb{L}$ with fixing other variables. 
In addition to the Lagrangian multipliers, there are 5 variables, including ${\bf Z}$, ${\bf Q}$, ${\bf E}$, ${\bf A}$ and ${\bf w}$, to be solved. 
The solutions of these subproblems are discussed below.

\subsubsection{Solving ${\bf Z}$}
With other variables in~\eqref{eq:graph_model1_ALM} fixed, the ${\bf Z}$-subproblem can be written as:
\begin{equation}
\label{eq:Z1}
\begin{aligned}
&\underset{{\bf Z}}{\min}~~\alpha||{\bf Z}||_1+f({\bf Z},{\bf Q}^k,{\bf E}^k,{\bf w}^k,{\bf Y}_1^k,{\bf Y}_2^k,\mu^k).
\end{aligned}
\end{equation}

To avoid using an auxiliary variable and matrix inversions, we use the linearized ADMM method~\cite{Lin11nips} to minimize the ${\bf Z}$-subproblem of~\eqref{eq:graph_model1_ALM}. 
The quadratic term $f$ is replaced by its first order approximation at the previous iteration 
and adding a proximal term. 
Thus, ${\bf Z}^{k+1}$ can be updated by:
\begin{equation}
\label{eq:Z}
\begin{aligned}
&\arg\underset{{\bf Z}}{\min}~~||{\bf Z}||_1+\frac{\eta\mu^k}{2\alpha}||{\bf Z}-{\bf Z}^k||_F^2+\langle\nabla_{\bf Z}f^k,{\bf Z}-{\bf Z}^k\rangle,
\end{aligned}
\end{equation}
where $f^k$ is the shorthand of $f({\bf Z}^k,{\bf Q}^k,{\bf E}^k,{\bf w}^k,{\bf Y}_1^k,{\bf Y}_2^k,\mu^k)$. 
In \eqref{eq:Z}, 
$\nabla_{\bf Z}f$ is the partial differential of $f$ with respect to ${\bf Z}$, and $\eta=||{\bf X}||_F^2$. 
With some manipulation,  we have: $\nabla_{\bf Z}f=-\mu({\bf X}^{\top}({\bf X}-{\bf XZ}-{\bf E}+{\bf Y}_1/\mu)-({\bf Z}-{\bf Q}+{\bf Y}_3/\mu))$.

Generally, the solution of ${\bf Z}^{k+1}$ is obtained by the soft-threshold (or shrinkage) method~\cite{Lin09report}:
\begin{equation}
\label{eq:updateZ}
\begin{aligned}
{\bf Z}^{k+1}=\mathbb{S}_{\frac{\alpha}{\eta\mu^k}}({\bf P}^k),
\end{aligned}
\end{equation}
where ${\bf P}^k={\bf Z}^k-\tau^k\nabla_{\bf Z}f^k\in\mathbb{R}^{n\times n}$, and $\mathbb{S}_{\frac{\alpha}{\eta\mu^k}}({\bf P}^k)$ is the soft-threshold operator on ${\bf P}^k$ with parameter $\frac{\alpha}{\eta\mu^k}$.

\subsubsection{Solving ${\bf Q}$}
By fixing other variables in~\eqref{eq:graph_model1_ALM}, the ${\bf Q}$-subproblem can be formulated as:
\begin{equation}
\label{eq:Q}
\begin{aligned}
&\underset{{\bf Q}}{\min}~~\gamma~\mbox{tr}({\bf Q}{\bf L}_{\bf A}{\bf Q}^{\top})+||{\bf Z}-{\bf Q}+{\bf Y}_2/\mu||_F^2.
\end{aligned}
\end{equation}

To compute ${\bf Q}$, we take the derivative of $\mathbb{L}$ with respect to ${\bf Q}$, and set it to be 0. 
With some manipulation, we have:
\begin{equation}
\label{eq:updateQ}
\begin{aligned}
&{\bf Q}^{k+1}=({\bf Z}^{k+1}+{\bf Y}_2^k/\mu^k)({\bf I}+\gamma({\bf L}_{{\bf A}^k}+{\bf L}_{{\bf A}^k}^{\top}))^{-1},
\end{aligned}
\end{equation}
where ${\bf I}$ is the identity matrix.

\subsubsection{Solving ${\bf E}$}
The ${\bf E}$-subproblem can be formulated as follows when
other variables in~\eqref{eq:graph_model1_ALM} are fixed:
\begin{equation}
\label{eq:E}
\begin{aligned}
&{\bf E}^{k+1}=\arg\underset{{\bf E}}{\min}~~||{\bf E}||_{2,1}+\frac{\mu^k}{2\beta}||{\bf X}-{\bf XZ}^{k+1}-{\bf E}+\frac{{\bf Y}_1^k}{\mu^k}||_F^2.
\end{aligned}
\end{equation}
which is computed by the $\ell_{2,1}$ minimization method~\cite{Liu13pami}:
\begin{equation}
\label{eq:updateE}
\begin{aligned}
{\bf E}^{k+1}=\mathcal{S}_{\frac{\beta}{\mu^k}}({\bf XZ}^{k+1}-{\bf X}-\frac{{\bf Y}_1^k}{\mu^k}),
\end{aligned}
\end{equation}
where $\mathcal{S}_{\frac{\beta}{\mu^k}}(\cdot)$ is the $\ell_{2,1}$ minimization operator with parameter $\frac{\beta}{\mu^k}$.

\subsubsection{Solving ${\bf A}$}
When other variables in~\eqref{eq:graph_model1_ALM} are fixed, the ${\bf A}$-subproblem can be formulated as:
\begin{equation}
\label{eq:A}
\begin{aligned}
&\underset{{\bf A1}={\bf 1},{\bf A}\geq 0}{\min}~~\gamma~\mbox{tr}({\bf Q}^{k+1}{\bf L}_{\bf A}{\bf Q}^{(k+1)T})\\
&+\lambda_1\sum_{i,j}{{\bf a}_{ij}({\bf w}_i^k-{\bf w}_j^k)^2}+\frac{\lambda_3}{2}||{\bf A}||_F^2.
\end{aligned}
\end{equation}

We separate~\eqref{eq:A} into a set of independent problems, and each ${\bf a}_i$ can be computed efficiently with a closed-form solution (please see the appendix of~\cite{Guo15ijcai} for details) as:
\begin{equation}
\label{eq:updateA}
\begin{aligned}
&{\bf a}_i^{k+1}=(\frac{1+\sum_{j=1}^\xi{\bf \hat{u}}_{ij}^{{\bf Q}^{k+1}}}{\xi}{\bf 1}-{\bf u}_i^{{\bf Q}^{k+1}})_+,
\end{aligned}
\end{equation}
where ${\bf u}_i^{{\bf Q}^{k+1}}\in\mathbb{R}^{n\times 1}$ is a vector whose $j$-th element is ${\bf u}_{ij}^{{\bf Q}^{k+1}}=\frac{\frac{\gamma}{2}||{\bf q}_i^{k+1}-{\bf q}_j^{k+1}||_F^2+\lambda_1({\bf w}_i^k-{\bf w}_j^k)^2}{\lambda_3}$. Notice that the parameter ${\xi\in\{1, \ldots,n\}}$ is introduced to control the number of nearest neighbors of ${\bf q}_i$ (or ${\bf x}_i$) that could have chance to connect edges with ${\bf q}_i$ (or ${\bf x}_i$).

\subsubsection{Solving ${\bf w}$}
By fixing other variables in~\eqref{eq:graph_model1_ALM}, the ${\bf w}$-subproblem can be formulated as:
\begin{equation}
\label{eq:w}
\begin{aligned}
&\underset{{\bf w}\geq 0}{\min}~~\lambda_1\sum_{i,j}{{\bf a}^{k+1}_{ij}({\bf w}_i-{\bf w}_j)^2}+\frac{\lambda_2}{2}||\Gamma\circ({\bf w}-{\bf r})||^2\\
&+\frac{\lambda_4}{2}||{\bf w}||^2.
\end{aligned}
\end{equation}
Similar to the solution for updating ${\bf Q}$, we take the derivative of $\mathbb{L}$ with respect to ${\bf w}$, and set it to be 0. 
With some manipulation, the closed-form solution of this subproblem can be computed by:
\begin{equation}
\label{eq:updateW}
\begin{aligned}
&{\bf w}^{k+1}=[(2\lambda_1({\bf D}^{k+1}-{\bf A}^{k+1}-{\bf A}^{(k+1)T})+\lambda_2{\Gamma}'\\
&+\lambda_4{\bf I})^{-1}(\lambda_2\Gamma\circ{\bf r})]_+,
\end{aligned}
\end{equation}
where ${\bf D}$ is the degree matrix of $({\bf A}+{\bf A}^{\top})$ that ${\bf D}=\mathop{\rm diag}\{{\bf d}_{11},{\bf d}_{22}, \ldots,{\bf d}_{nn}\}$, 
and ${\bf d}_{ii}=\sum_j({\bf a}_{ij}+{\bf a}_{ji})$, and ${\Gamma}'=\mathop{\rm diag}\{\Gamma_1,\Gamma_2, \ldots,\Gamma_n\}$.

%
The procedure of solving~\eqref{eq:graph_model1_ALM} terminates when the maximum element changes of ${\bf Z}$, ${\bf Q}$, ${\bf E}$, ${\bf A}$ and ${\bf w}$ between two consecutive iterations are less than 
a threshold (e.g., $10^{-6}$ in this work) or the maximum number of iterations reaches 
a pre-defined number (e.g., $100$ in this work).

\begin{algorithm}[t]
\caption{Optimization Procedure to~\eqref{eq:graph_model1_separable}}\label{alg::optimization}
\begin{algorithmic}[1]
\REQUIRE
The patch feature matrix ${\bf X}$ and the initial weight vector ${\bf r}$, the parameters $\alpha$, $\beta$, $\gamma$, $\lambda_1$, $\lambda_2$, $\lambda_3$ and $\lambda_4$;\\
Set ${\bf Z}_0={\bf Q}_0={\bf A}_0={\bf Y}_{2,0}={\bf 0}$, ${\bf E}_0={\bf Y}_{1,0}={\bf 0}$, ${\bf w}={\bf 1}$, $\mu_0=0.1$, $\mu_{max}=10^{10}$, $\rho=1.1$ and $k=0$.
\ENSURE
${\bf Z}$, ${\bf Q}$, ${\bf E}$, ${\bf A}$ and ${\bf w}$.
\WHILE{not converged}
\STATE Update ${\bf Z}^{k+1}$ by~\eqref{eq:updateZ};
\STATE Update ${\bf Q}^{k+1}$ by~\eqref{eq:updateQ};
\STATE Update ${\bf E}^{k+1}$ by~\eqref{eq:updateE};
\FOR{$i$ from $1$ to $n$}
\STATE Update ${\bf a}_i^{k+1}$ by~\eqref{eq:updateA};
\ENDFOR
\STATE Update ${\bf w}^{k+1}$ by~\eqref{eq:updateW};
\STATE Update Lagrange multipliers;
\STATE Update $\mu^{k+1}$ by $\mu^{k+1}=\min(\mu_{max},\rho\mu^k);$
\STATE Update $k$ by $k=k+1$.
\ENDWHILE
\end{algorithmic}
\end{algorithm}

Algorithm~\ref{alg::optimization} summarizes the optimization procedures. 
Since each subproblem of~\eqref{eq:graph_model1_ALM} is convex,  
the solution by the proposed algorithm satisfies the Nash equilibrium conditions~\cite{Xu13siam}.

\section{Structured SVM Tracking}
\label{sec::tracking}

In this section, we incorporate the optimized weights of patches 
into the conventional tracking-by-detection algorithm, Struck~\cite{Stuck11iccv}, for visual tracking. 
Although we use the Struck method in this work, the optimized patch weights can also be incorporated into other tracking-by-detection algorithms. 
The Struck method selects the optimal target bounding box $b^*_t$ in the $t$-th frame by maximizing a classification score:
\begin{equation}
\label{eq:struck}
\begin{aligned}
&b^*_t=\arg\underset{b}{\max}~\langle{\bf h}_{t-1},{\bf x}_{t,b}\rangle,
\end{aligned}
\end{equation}
where ${\bf h}_{t-1}$ is the normal vector of a decision plane of the $(t-1)$-th frame, 
and ${\bf x}_{t,b}=[{\bf x}_{t,1};{\bf x}_{t,2}; \ldots;{\bf x}_{t,n}]$ denotes the descriptor representing a bounding box $b$ in the $t$-th frame. 
Instead of using binary-labeled samples, the Struck method employs structured samples that 
consist of a target bounding box and nearby boxes in the same frame to alleviate the labeling ambiguity in training the classifier. 
Specifically, it enforces that the confidence score of a target bounding box is larger than that of a nearby box by a margin determined by the overlap ratio between two boxes. 
As such, the Struck method can reduce adverse the labeling ambiguity problems.

For robust tracking, we decompose the problem of target state estimation into the two subproblems of translation estimation and scale estimation~\cite{Ma15cvpr,MuSTer15cvpr,ScaleCSK14bmvc}. 
Motivated by Bayesian filtering algorithms~\cite{Li16tip1,MCPF17cvpr}, we propose a simpler yet effective random strategy for target state refinement.

\subsection{Translation Estimation}
We incorporate the optimized patch weights into the Struck method, in which 
we improve the robustness to drastic appearance changes 
and unreliable tracking results of a target object. 
Given the bounding box of the target object in the previous frame $t-1$, we first set a searching window in current frame $t$. 
For $i$-th candidate bounding box within the search window, we weight its patch feature descriptor ${\bf x}_{t,i}$ by the weight $\hat{\bf w}_{t-1,i}=1/(1+\exp(-\sigma\overline{\bf w}_{t-1,i}))$, and concatenate them into a vector as the feature representation:
\begin{equation}
\label{eq:improved_feature}
\begin{aligned}
\hat{\bf x}=[\hat{\bf w}_{t-1,1}{\bf x}_{t,1};\hat{\bf w}_{t-1,2}{\bf x}_{t,2}; \ldots;\hat{\bf w}_{t-1,n}{\bf x}_{t,n}],
\end{aligned}
\end{equation}
{where we normalize ${\bf w}$ as $\overline{\bf w}$ so that all elements of $\overline{\bf w}$ sum to 1,} and the parameter $\sigma$ is fixed to a pre-defined number (e.g., 37 in this work). 
{Herein, we use the Sigmoid function to map the normalized weights into the range of 0 to 1, which has a parameter $\sigma$ to control the steepness of normalized weights.}
The optimal bounding box $b^*_t$ can be selected to update the object location by maximizing the classification score:
\begin{equation}
\label{eq:improved_struck}
\begin{aligned}
&{\hat b}_t=\arg\underset{b}{\max}~(\omega\langle{\bf h}_{t-1},\hat{\bf x}_{t,b}\rangle+(1-\omega)\langle{\bf h}_0,\hat{\bf x}_{t,b}\rangle),
\end{aligned}
\end{equation}
where ${\bf h}_0$ is learned in the initial frame, which can alleviate the issue 
of learning drastic appearance changes, and $\omega$ is a weight parameter.

\subsection{Scale Estimation}
%
%

Given the estimated location ${\hat b}_t$, we sample a set $\mathbb{B}_t$ of bounding boxes from the Gaussian distribution centered at ${\hat b}_t$, in which the elements of the covariance are the variations of the affine parameters, and its setting depends on motion variations of the target object.
To simultaneously estimate scales and refine locations, we utilize four independent affine parameters to draw samples including the scale factor, aspect ratio and translation.
%
For example, we empirically set to these parameters (scale factor, aspect ratio, and translation) to 0.05, 0.01, 1 and 1, respectively in this paper.
{As object translation is estimated before, we use 100 samples in this paper to compute scale while slightly adjusting translation for a trade-off between efficiency and accuracy.}
The bounding box is updated by the one with the highest score,
\begin{equation}
\label{eq:scaler}
\begin{aligned}
&b^*_t=\arg\underset{b\in \mathbb{B}_t}{\max}~(\omega\langle{\bf g}_{t-1},\hat{\bf x}_{t,b}\rangle+(1-\omega)\langle{\bf g}_0,\hat{\bf x}_{t,b}\rangle),
\end{aligned}
\end{equation}
where ${\bf g}_{t-1}$ and ${\bf g}_0$ are classifiers trained in scale spaces at time $t-1$ and 0, respectively. 

To update the classifier ${\bf g}_t$, we use a similar method to translation estimation. 
%
Given the optimal estimate $b^*_t$, we extract bounding boxes $\mathbb{S}_t$ around $b^*_t$ at different scales and the corresponding feature representations for scale factors $\{0.50, 0.52, \ldots,0.98,1.02, \ldots,1.48,1.50\}$ excluding the positive sample with the scale factor 1~\cite{SOWP17tip}.
We then find the optimal ${\bf g}^*_t$ by
\begin{equation}
\label{eq:scaler_update}
\begin{aligned}
&{\bf g}^*_t=\arg\underset{{\bf g}_t}{\min}~\xi||{\bf g}_t||^2+\sum_{b\in\mathbb{S}_t}\max\{0,\triangle(b,b_t^*)-\langle{\bf g}_t,{\bf x}_b-{\bf x}_{b_t^*}\rangle\},
\end{aligned}
\end{equation}
where $\triangle(b_t,b)=1-IoU(b_t,b)$, and $IoU$ indicates the Intersection-over-Union operation.
 To optimize~\eqref{eq:scaler_update}, we use the stochastic variance reduced gradient scheme~\cite{Johnson13nips}. 
 To reduce the sensitivity to noises of scale update, we carry out scale estimation at an interval of 3 frames.

\subsection{Model Update}
To alleviate the issues of model drift, we update the classifier and patch weights only when the confidence score of tracking result is larger than a threshold $\theta$.
 In this paper, the confidence score of tracking result in $t$-th frame is defined as the average similarity between the weighted descriptor of the tracked bounding box and the positive support vectors
 \begin{equation}
 \frac{1}{|\mathbb{P}_t|}\sum_{{\bf s}\in\mathbb{P}_t}\langle{\bf s},\hat{\bf x}_{t,b^*_t}\rangle,
 \end{equation}
 where $\mathbb{P}_t$ is the set of the positive support vectors at time $t$. 
Algorithm~\ref{alg::tracking} shows the main steps of the proposed tracking method.

\begin{algorithm}[t]
\caption{Proposed Object Tracking Algorithm}\label{alg::tracking}
\begin{algorithmic}[1]
\REQUIRE
Input video sequence,\\
initial target bounding box $b_0$.
\ENSURE
Estimated object state $b^*_t$.
\STATE {\tt // Initialization}
\STATE Compute $\hat{\bf w}_0$, ${\bf h}_0$ and ${\bf g}_0$ according to $b_0$;
\REPEAT
\STATE Set the searching window in $t$-th frame according to $b^*_{t-1}$ and extract features ${\bf x}_t$;
\STATE {\tt // Feature construction}
\STATE Construct feature representation $\hat{\bf x}_t$ using ${\bf x}_t$ and $\hat{\bf w}_{t-1}$;
\STATE {\tt // Translation estimation}
\STATE Estimate object location ${\hat b}_t$ by~\eqref{eq:improved_struck};
\STATE {\tt // Scale estimation}
\STATE Estimate final object state $b^*_t$ by~\eqref{eq:scaler};
\STATE {\tt // Model and weight update}
\IF{$\frac{1}{|\mathbb{P}_t|}\sum_{{\bf s}\in\mathbb{P}_t}\langle{\bf s},\hat{\bf x}_{t,b^*_t}\rangle<\theta$}
\STATE {\tt // Weight computation}
\STATE Run Algorithm~\ref{alg::optimization} for computing the patch weights $\hat{\bf w}_t$ according to $b^*_t$;
\STATE Update ${\bf h}_t$ and ${\bf g}_t$;
\ENDIF
\UNTIL{\emph{End of video sequence}.}
\end{algorithmic}
\end{algorithm}

\subsection{Discussion}
\label{sec::difference}
It should be noted that the proposed tracking algorithm is significantly different from the recently proposed approaches that use sparse representation for object tracking~\cite{Mei11pami,SCM12cvpr,Bao12cvpr,Li16tip1} in which  reconstruction errors or representation coefficients are used to compute the confidence of candidates in the Bayesian filtering framework. 
While we employ the sparse representation to learn a dynamic graph for representing a target object, the node weights are used to suppress 
the effects of background clutter in the tracking-by-detection framework.

In addition, our approach is also significantly different from the SOWP~\cite{Kim15iccv} method in several aspects.
First, the proposed algorithm learns a dynamic graph to represent a target object that better captures the intrinsic relationship among image patches. 
Second, our method optimizes the edge and node weights jointly while the SOWP method first computes the edge weights and then the node weights. 
Third, the proposed tracker considers the initial foreground and background clutter in a unified model, while the SOWP method requires two passes to
compute the final patch weights, one for foreground and another for background. 
%
%



\section{Performance Evaluation}
\label{sec::experiments}
{The proposed tracker based weighted patch-based graph (WPG) representation is  
implemented in C++.
All experiments are carried out on a machine with an Intel i7 4.0 GHz CPU and 32 GB RAM.
%
We test runtime of WPG on the OTB100 dataset~\cite{RGBbenchmark15pami}, and scale each frame such that the minimum side length of a bounding box is 32 pixels for efficiency. 
The proposed algorithm is able to track a target object at 5 frames per second where the optimization method converges within 50 iterations.}
We use the benchmark datasets and protocols~\cite{RGBbenchmark15pami,NUS-PRO16pami,vot14challenge} to evaluate the proposed approach. 
In addition, we evaluate several variants of the proposed method to demonstrate the contribution of main modules.

\subsection{Experimental Setup}
\subsubsection{Parameters}
For fair comparisons, we fix all parameters and other settings on all datasets in our experiments.
We partition all bounding box into 64 non-overlapping patches as a trade-off between accuracy and efficiency, and extract color and gradient histograms for each patch, where the dimension of gradients and each color channel is set to be 8.
%
%
We  evaluate different number of patches from $\{36, 49, 64, 81, 100\}$, and empirically determine that the proposed method performs best
with 64  patches as a trade-off between accuracy and complexity. 
{Note that we fix patch number as square to adapt patch size to the size of object bounding box, which makes patches have consistent shape with target object. Otherwise, it is hard to find a unified partition method for all sequences.}
To improve efficiency, each frame is scaled such that the minimum side length of a bounding box is 32 pixels. 
A bounding box is described by with and height of $W$ and $H$ pixels.
The side length of a search window is initially set to a small range, $0.8\sqrt{WH}$, to reduce false positives, and then set to a large range, $\sqrt{WH}$, to handle abrupt motions if the center distance of the object box between two consecutive frames is above a predefined threshold (e.g., 5 pixels in this work). 

\begin{figure}[t]
\centering
\includegraphics[width =\columnwidth]{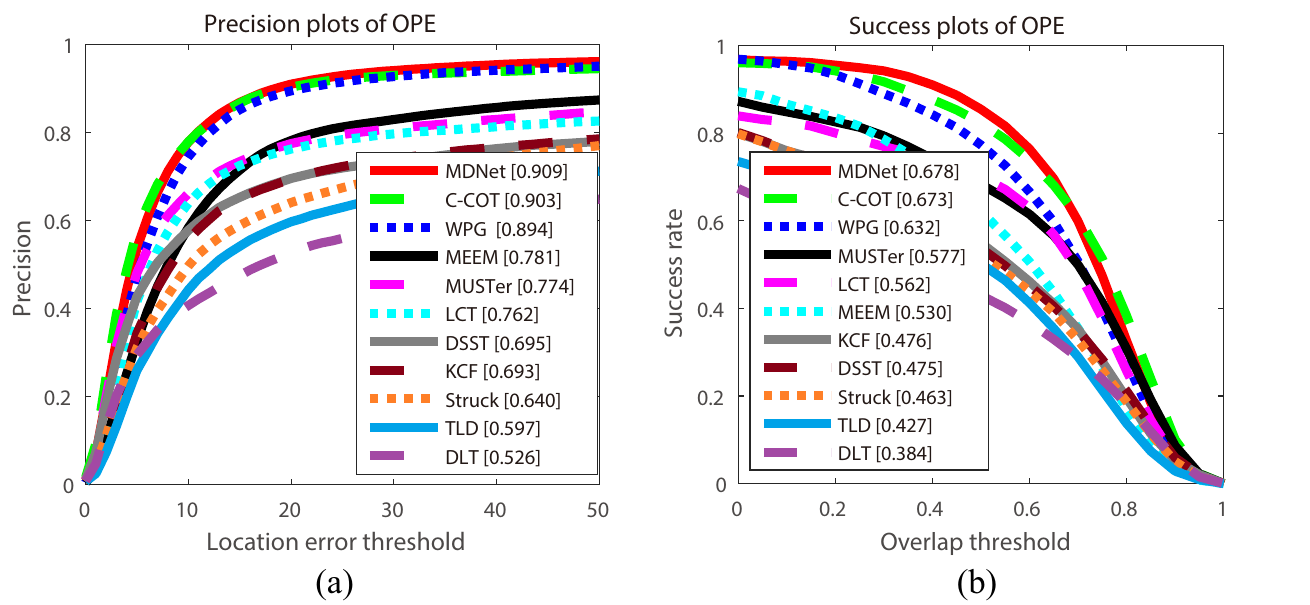}
\caption{
 Precision and success plots of OPE (one-pass evaluation)~\cite{RGBbenchmark15pami} of the proposed tracker against other state-of-the-art trackers on OTB100. The representative score of PR is presented in the legend. }\label{fig::OTB100}
\end{figure} 

\begin{figure*}[t]
\centering
\includegraphics[width =1.8\columnwidth]{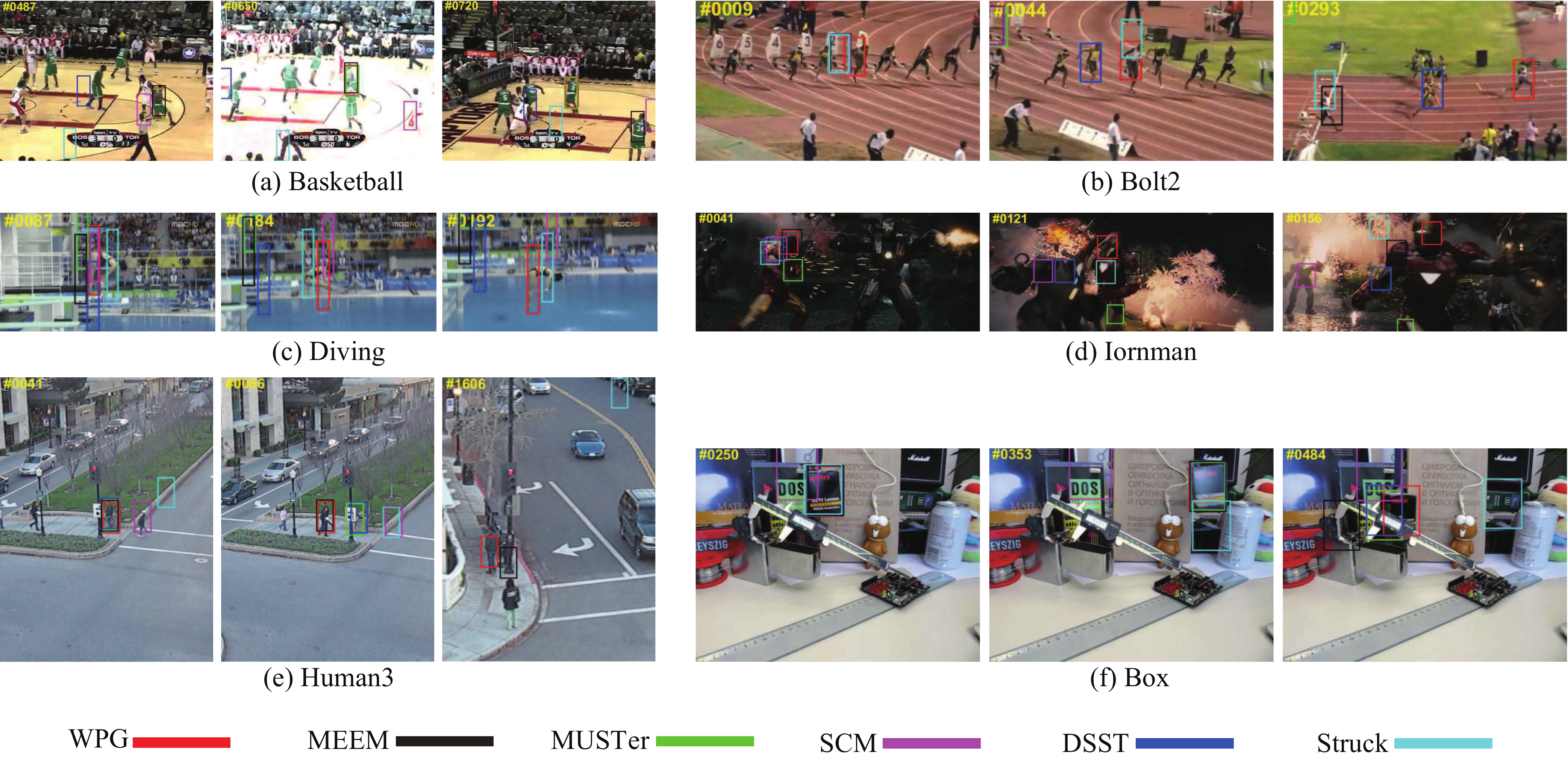}
\caption{
Sample results of our method against Struck~\cite{Stuck11iccv}, MEEM~\cite{MEEM14eccv}, MUSTer~\cite{MuSTer15cvpr}, DSST~\cite{ScaleCSK14bmvc} and SCM~\cite{SCM12cvpr}.}\label{fig::sample_results}
\end{figure*}

For seed selection, we shrink and expand the tracked bounding box $(lx, ly, W, H)$ as $(lx+0.2W,  ly+0.2H,  0.6W,  0.6H)$ and $(lx-W', ly-H', W+2W', H+2H')$, respectively, where $(lx, ly)$ denotes the upper left coordinate of the tracked bounding box, and $W'$ and $H'$ indicate the patch width and height, respectively~\cite{Kim15iccv}. 
%
In the proposed model~\eqref{eq:graph_model1}, there involves several parameters, which are set as follows. On one hand, similar to~\cite{Guo15ijcai}, we simplify the settings as $\alpha=\beta$ and $\lambda_3=\lambda_4$. Following~\cite{Guo15ijcai}, we set $\{\alpha, \beta, \gamma, \lambda_3, \lambda_4, \xi\}=\{0.1, 0.1, 10, 1, 1, 6\}$. {Although $\gamma$ is 2 orders of magnitude higher than $\alpha$ and $\beta$, we find that these terms can balance well by outputting each term after optimization.} On the other hand, $\lambda_1$ and $\lambda_2$ are to control the balance of smoothness and fitness of ${\bf w}$. According to the setting of similar models~\cite{Yang13cvpr,Li17aaai}, we set $\{\lambda_1, \lambda_2\}=\{5, 0.5\}$. 
For the Struck method, we empirically set $\{\omega,\theta\}=\{0.67,0.25\}$~\cite{Kim15iccv}. 

\subsubsection{OTB100  Dataset}
We evaluate the proposed tracking method on the OTB100 benchmark dataset~\cite{RGBbenchmark15pami}. 
The OTB100 dataset contains $100$ image sequences with ground-truth object locations and attributes for performance analysis. 
We use precision rate (PR) and success rate (SR) with the threshold of $20$ pixels for quantitative performance. 
%

\subsubsection{Temple Color  Dataset}
For further validating the effectiveness of the proposed approach, we also compare with other
tracking approaches on the Temple Color dataset~\cite{Temple-Color15tip}.
This database contains 128 challenging image sequences of human, animals and rigid objects. 
In addition to tracking ground truth, each sequence in the dataset is also annotated by its challenging factors as defined in~\cite{RGBbenchmark15pami}. 
The evaluation metrics are also defined in~\cite{RGBbenchmark15pami}.

\subsubsection{NUS PRO Dataset}
We also compare the proposed algorithm with other tracking approaches on the NUS-PRO dataset~\cite{NUS-PRO16pami}.
This dataset contains $365$ challenging image sequences of pedestrians and rigid objects, mainly captured from moving cameras.
Aside from target locations, each sequence is annotated with occlusion level for evaluation. 
 We employ the threshold-response relationship (TRR) with three criteria for occlusion computation~\cite{NUS-PRO16pami}
 on the entire dataset to evaluate the proposed tracking method.

\begin{table*}[t]
\caption{Attribute-based PR scores on OTB benchmark compared with recent trackers, where the best results of deep and non-deep trackers divided by dash line are in \textcolor{red}{red} and \textcolor{green}{green} colors, respectively. }
\centering
\begin{tabular}{c||c c c c : c c c c c c c c}
	\hline
	   & MDNet & C-COT & HCF & DLT & SOWP & MEEM & MUSTer & KCF & LCT & DSST & Struck & WPG\\
   \hline
   IV & \textcolor{red}{0.911}&0.878 &0.817 & 0.522 & 0.777 & 0.740 & 0.782 & 0.708 & 0.746 & 0.723 & 0.545 & \textcolor{green}{0.873} \\
	SV &\textcolor{red}{0.892} &0.881 &0.802 & 0.542 & 0.750 & 0.740 & 0.715 & 0.639 & 0.686 & 0.667 & 0.600 & \textcolor{green}{0.858} \\
	OCC &0.857 & \textcolor{red}{0.904}&0.767 & 0.454 & 0.754 & 0.741 & 0.734 & 0.622 & 0.682 & 0.615 & 0.537 & \textcolor{green}{0.863} \\
	DEF &\textcolor{red}{0.899} &0.865 &0.791 & 0.451 & 0.741 & 0.754 & 0.689 & 0.617 & 0.689 & 0.568 & 0.527 & \textcolor{green}{0.878} \\
	MB &0.866 &\textcolor{red}{0.899} &0.797 & 0.427 & 0.710 & 0.722 & 0.699 & 0.617 & 0.673 & 0.636 & 0.594 & \textcolor{green}{0.817} \\
	FM &\textcolor{red}{0.885} & 0.883&0.797 & 0.426 & 0.719 & 0.735 & 0.691 & 0.628 & 0.675 & 0.602 & 0.626 & \textcolor{green}{0.824} \\
	IPR &\textcolor{red}{0.910} &0.877 &0.854 & 0.471 & 0.828 & 0.794 & 0.773 & 0.693 & 0.782 & 0.724 & 0.637 & \textcolor{green}{0.877} \\
	OPR &\textcolor{red}{0.900} &0.899 &0.810 & 0.517 & 0.790 & 0.798 & 0.748 & 0.675 & 0.750 & 0.675 & 0.593 & \textcolor{green}{0.882} \\
	OV & 0.825& \textcolor{red}{0.895}&0.677 & 0.558 & 0.633 & 0.685 & 0.591 & 0.498 & 0.558 & 0.487 & 0.503 & \textcolor{green}{0.802} \\
	BC & \textcolor{red}{0.925}&0.882 &0.847 & 0.509 & 0.781 & 0.752 & 0.786 & 0.716 & 0.740 & 0.708 & 0.566 & \textcolor{green}{0.885} \\
	LR &0.942 &\textcolor{red}{0.975} &0.787 & 0.615 & 0.713 & 0.605 & 0.677 & 0.545 & 0.490 & 0.595 & 0.674 & \textcolor{green}{0.948} \\
	\hline
	All & \textcolor{red}{0.909}& 0.903& 0.837 & 0.526 & 0.803 & 0.781 & 0.774 & 0.692 & 0.762 & 0.695 & 0.640 & \textcolor{green}{0.894} \\
	\hline
\end{tabular}
\label{tb::attribute_pr}
\end{table*}

\begin{table*}[t]
\caption{Attribute-based SR scores on OTB benchmark compared with recent trackers, where the best results of deep and non-deep trackers divided by dash line are in \textcolor{red}{red} and \textcolor{green}{green} colors, respectively. }
\centering
\begin{tabular}{c||c c c c : c c c c c c c c}
	\hline
	   & MDNet & C-COT & HCF & DLT & SOWP & MEEM & MUSTer & KCF & LCT & DSST & Struck & WPG\\
   \hline
   IV & \textcolor{red}{0.684}&0.674 & 0.540 & 0.408 & 0.554 & 0.517 & 0.600 & 0.474 & 0.566 & 0.489 & 0.422 & \textcolor{green}{0.632} \\
	SV &\textcolor{red}{0.658} &0.654 & 0.488 & 0.399 & 0.478 & 0.474 & 0.518 & 0.399 & 0.492 & 0.413 & 0.404 & \textcolor{green}{0.579} \\
	OCC &0.646 &\textcolor{red}{0.674} & 0.525 & 0.335 & 0.528 & 0.504 & 0.554 & 0.438 & /0.507 & 0.426 & 0.394 & \textcolor{green}{0.619} \\
	DEF &\textcolor{red}{0.649} & 0.614& 0.530 & 0.295 & 0.527 & 0.489 & 0.524 & 0.436 & 0.499 & 0.412 & 0.383 & \textcolor{green}{0.605} \\
	MB & 0.679&\textcolor{red}{0.706} & 0.573 & 0.353& 0.557 & 0.545 & 0.557 & 0.456 & 0.532 & 0.465 & 0.468 & \textcolor{green}{0.622} \\
	FM & 0.675&\textcolor{red}{0.676} & 0.555 & 0.345 & 0.542 & 0.529 & 0.539 & 0.455 & 0.527 & 0.440 & 0.470 & \textcolor{green}{0.603} \\
	IPR & \textcolor{red}{0.655}& 0.627& 0.559 & 0.348 & 0.567 & 0.529 & 0.551 & 0.465 & 0.557 & 0.485 & 0.453 & \textcolor{green}{0.606} \\
	OPR & \textcolor{red}{0.661}& 0.652& 0.537 & 0.376 & 0.549 & 0.528 & 0.541 & 0.454 & 0.541 & 0.453 & 0.424 & \textcolor{green}{0.608} \\
	OV &0.627 &\textcolor{red}{0.648} & 0.474 & 0.384 & 0.497 & 0.488 & 0.469 & 0.393 & 0.452 & 0.374 & 0.384 & \textcolor{green}{0.586} \\
	BC &\textcolor{red}{0.676} &0.652 & 0.587 & 0.553 & 0.575 & 0.523 & 0.579 & 0.498 & 0.481 & 0.373 & 0.438 & \textcolor{green}{0.652} \\
	LR & \textcolor{red}{0.631}& 0.629& 0.424 & 0.422 & 0.416 & 0.355 & 0.477 & 0.306 & 0.330 & 0.311 & 0.313 & \textcolor{green}{0.575} \\
	\hline
	All & \textcolor{red}{0.678}&0.673 & 0.562 & 0.384 & 0.560 & 0.530 & 0.577 & 0.475 & 0.562 & 0.475 & 0.463 & \textcolor{green}{0.632} \\
	\hline
\end{tabular}
\label{tb::attribute_sr}
\end{table*}

\begin{figure}[t]
\centering
\includegraphics[width = 0.8\columnwidth]{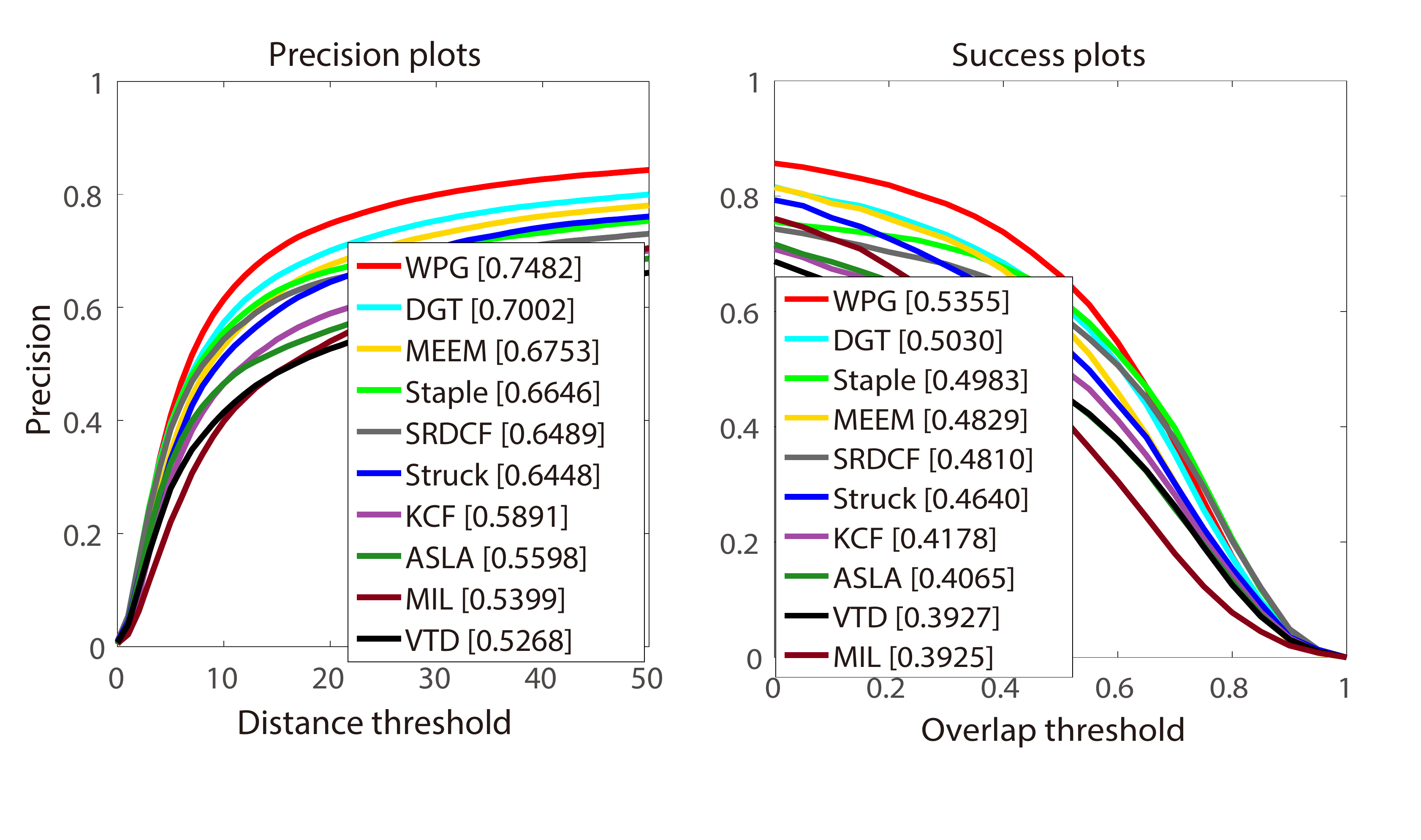}
\caption{
PR and SR curves on the Temple Color dataset where ten trackers are shown here.}\label{fig::Temple-Color_plots}
\end{figure}

\begin{figure*}[t]
\centering
\includegraphics[width = 1.6\columnwidth]{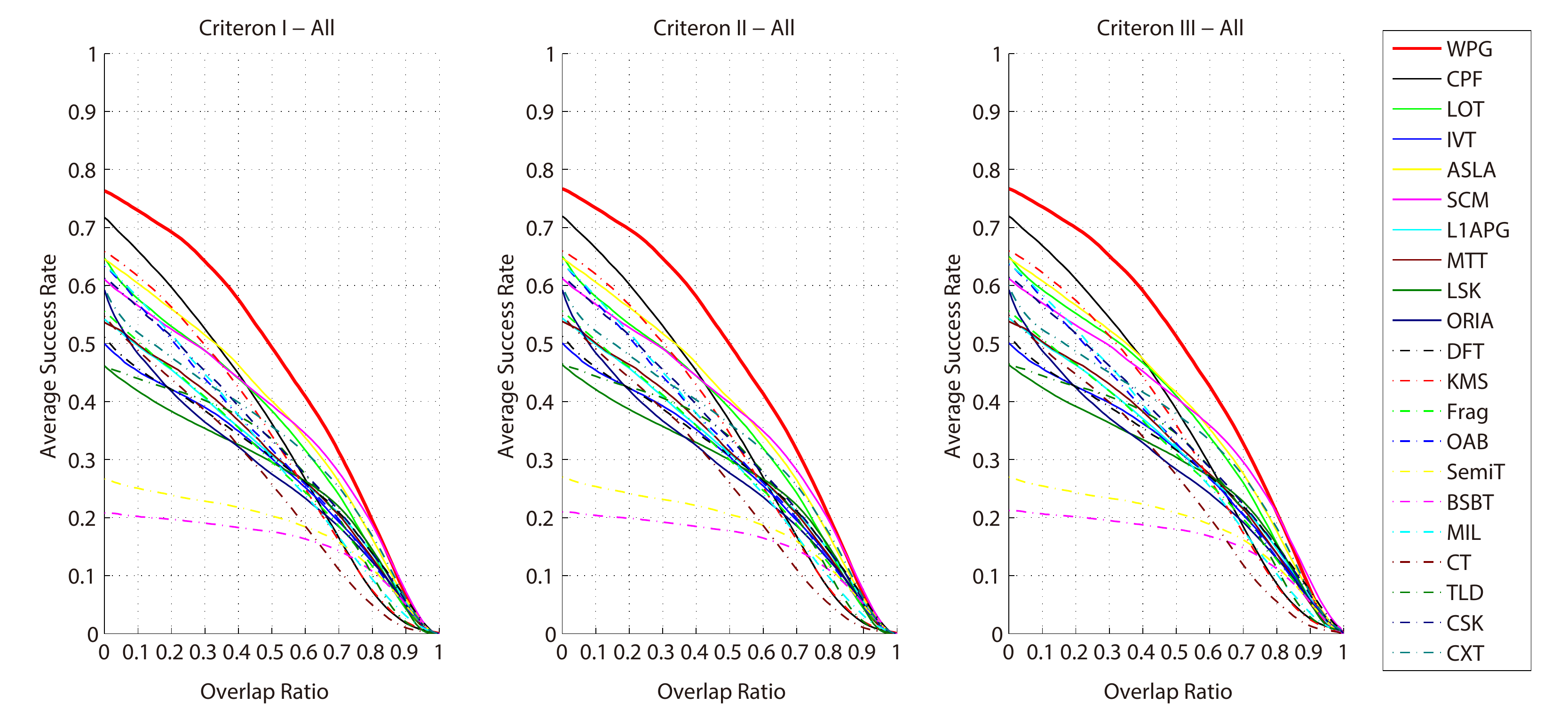}
\caption{
 TRR curves on NUS-PRO, where twenty trackers are shown here.}\label{fig::NUS-PRO_plots}
\end{figure*}

\begin{figure*}[t]
\centering
\includegraphics[width = 1.6\columnwidth]{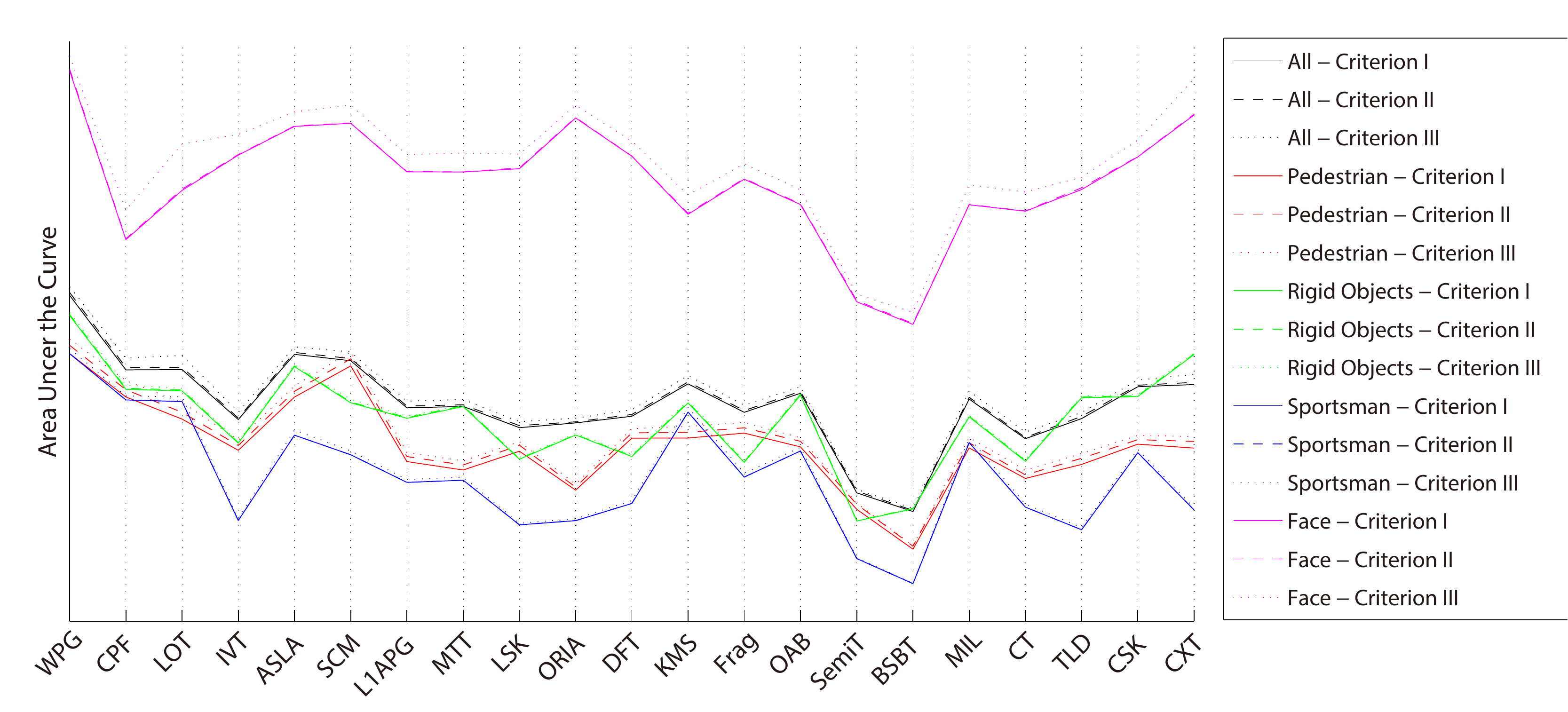}
\caption{
AUC plots of TRR curves with different object categories on NUS-PRO, where twenty trackers are shown here.}\label{fig::NUS-PRO_category}
\end{figure*}

\begin{figure*}[t]
\centering
\includegraphics[width = 1.6\columnwidth]{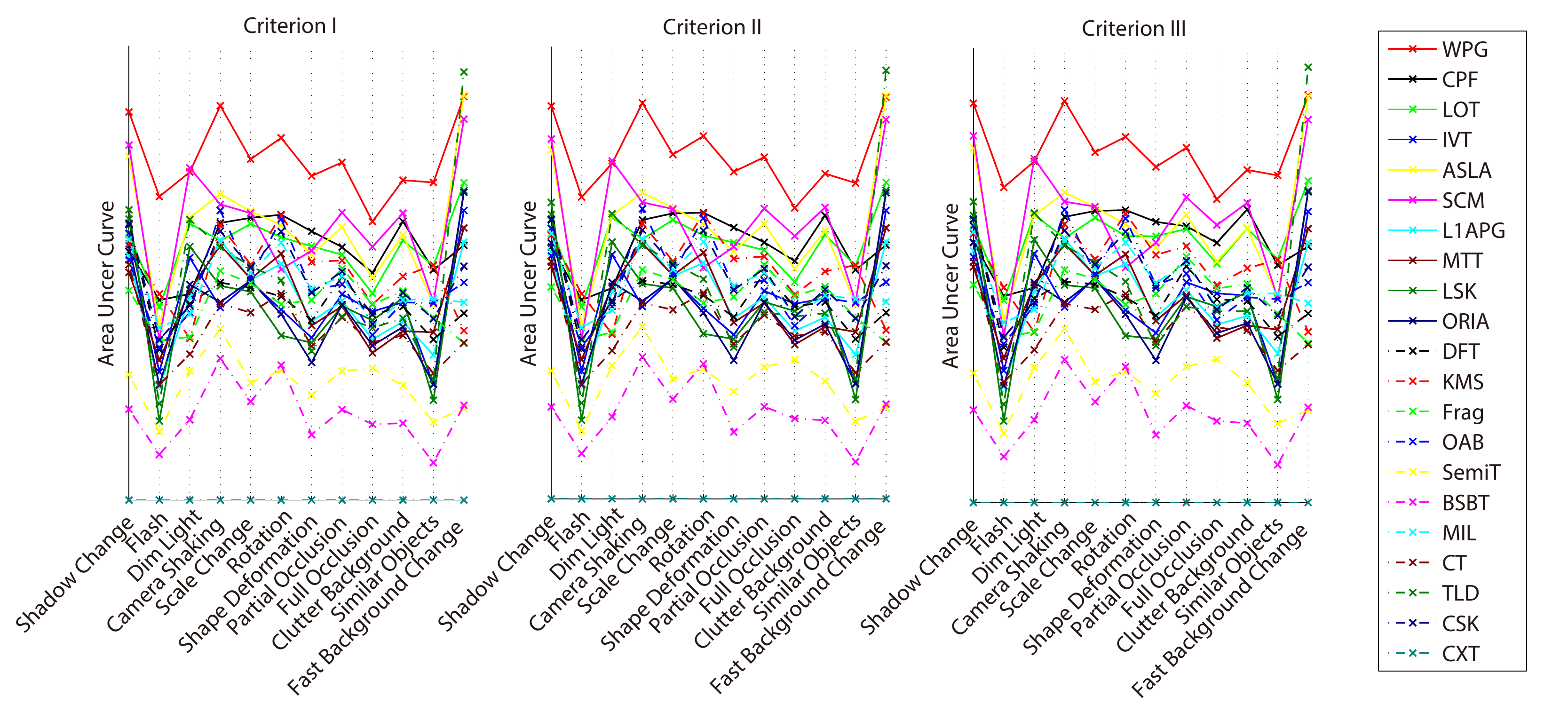}
\caption{
AUC plots of TRR curves with different challenges on NUS-PRO, where twenty trackers are shown here.}\label{fig::NUS-PRO_challenges}
\end{figure*}

{\subsubsection{VOT Challenge Dataset}
For more comprehensive evaluation, we also run the proposed tracker on the VOT2014 challenge dataset~\cite{vot14challenge}, whose dataset contains more deformations and the aligned bounding boxes contain more noise. 
Accuracy (ACC) and robustness (ROB) are used to assess the performance of a tracker. The accuracy computes the overlap ratio between an estimated bounding box and the ground truth. 
The robustness indicates the number of tracking failures, i.e., the number of frames in which the overlap ratios are zero.}

\subsection{Evaluation on the OTB100  Dataset}
We first evaluate the proposed algorithm on the OTB100 dataset against tracking methods.
Next we analyze the performance of evaluated methods based on attributes of image sequences.

\subsubsection{Tracking Methods Without Deep Features}
We evaluate the proposed algorithm against the state-of-the-art tracking methods without using deep features, e.g., Struck~\cite{Stuck11iccv}, DSST~\cite{ScaleCSK14bmvc}, MEEM~\cite{MEEM14eccv}, MUSTer~\cite{MuSTer15cvpr} and SOWP~\cite{Kim15iccv}. 
Figure~\ref{fig::OTB100} shows the OPE plots on the OTB100 dataset, and Figure~\ref{fig::sample_results} presents some qualitative results. 
Overall, the proposed algorithm performs favorably against the state-of-the-art methods, e.g.,  9.1\% over SOWP in the precision score and 5.5\% over MUSTer in the success score.
Figure~\ref{fig::sample_results} shows that the proposed approach effectively  handles scenes with illumination variation (\emph{Basketball} and \emph{Ironman}), background clutter (\emph{Diving}, \emph{Ironman} and \emph{Box}), deformation (\emph{Basketball}, \emph{Bolt2} and \emph{Diving}) and partial occlusion (\emph{Ironman}, \emph{Box} and \emph{Human3}).

{The excellent performance of WPG suggests that the proposed tracker is able to mitigate outlier effects by integrate local patch weights into feature representations, which brings biggest performance gain for achieving state-of-the-art tracking performance. In addition to it, the following components are also beneficial to promoting tracking performance. First, local patch representations are robust to object deformation and partial occlusion. Second, the classification and update schemes are used to avoid model contagious by drastic appearance changes and unreliable tracking results of a target object. Finally, the scale handling strategy is employed to adapt to scale variations and also refine object translation.}

\subsubsection{Tracking Methods Based on Deep Features}
We evaluate the proposed algorithm against the state-of-the-art tracking methods using deep features including DLT~\cite{Wang13nips}, HCF~\cite{Ma15iccv}, C-COT~\cite{C-COT16eccv} and MDNet~\cite{MDNet15cvpr}. 
Figure~\ref{fig::OTB100}, Table~\ref{tb::attribute_pr} and Table~\ref{tb::attribute_sr} show the evaluation results. 
Overall, the proposed tracker performs well against the DLT and HCF methods in all aspects.
The proposed tracker performs equally well against the C-COT and MDNet schemes in terms of precision and slightly worse in terms of success rate. 
Furthermore, the proposed algorithm differs from the C-COT and MDNet methods in several aspects.  
\begin{itemize}
\item 
The proposed tracking method does not require laborious pre-training or a large training set. 
In addition, it does not need to save a large pre-trained deep model.
We initialize the proposed model using the ground truth bounding box in the first frame, and update it in subsequent frames. 
\item It is easy to implement as each subproblem of the proposed model has a closed-form solution. 
%
\item It performs more robustly than the MDNet and C-COT methods in some situations. 
In particular, it outperforms the C-COT method on sequences with background clutters
in terms of precision and success rate,  which suggests the effectiveness of our approach in suppressing the background effects during tracking.
\end{itemize}

\subsubsection{Attribute-based Evaluation}
We present the precision plots with 11 different attributes in Table~\ref{tb::attribute_pr} and Table~\ref{tb::attribute_sr}. The attributes include background clutter (BC), deformation (DEF), fast motion (FM), illumination variation (IV), in-plane rotation (IPR), low resolution (LR), motion blur (MB), occlusion (OCC), out-of-plane rotation (OPR), out of view (OV) and scale variation (SV). 

The comparison plots show that our tracker significantly outperforms other non-DL-based tracking methods, and achieves comparable performance with DL-based ones on the attribute-based subsets (e.g., BC and DEF), which validates the effectiveness of introducing the optimized weights in the object representation that suppresses background clutter and noises. The performance of our tracker against others on OCC and OV suggests that the adopted classification and update schemes can re-track objects in case of tracking failure, e.g., totally occlusion and re-entering the field of view, and alleviate incorrect update of noisy samples. The even worse performance of our tracker against others on FM and LR suggests the weakness of our used features (color and gradient) in representing the target object and search strategy, and we will address these issues in future work.

{In particular, we compare our WPG with the SOWP method~\cite{Kim15iccv} that is most related to us as follows. For the PR score, WPG outperforms the SOWP method significantly, especially on the sequences with deformation, out of view, background clutter and low resolution. It demonstrates advances of WPG over SOWP in learning robust object feature representations under background inclusion and less information, and also in re-tracking objects after they back to view. For the SR score, WPG also excels SOWP with a large margin, especially on the challenges of scale variation, occlusion and low resolution, which verify the effectiveness of scale handling, background suppression and reliability highlighting in WPG, while SOWP does not handle scale variations and is also limited by its weight computation scheme.}

\subsection{Evaluation on the Temple Color Dataset}
We evaluate the proposed algorithm on the Temple Color dataset~\cite{Temple-Color15tip}. 
Figure~\ref{fig::Temple-Color_plots} shows the evaluation results against 9 state-of-the-art tracking approaches, including DGT~\cite{Li17aaai}, Staple~\cite{bertinetto2016staple}, MEEM~\cite{MEEM14eccv}, SRDCF~\cite{danelljan2015learning}, Struck~\cite{hare2016struck}, KCF~\cite{CSK15pami}, ASLA~\cite{Jia12cvpr}, MIL~\cite{MIT11pami}, and VTD~\cite{kwon2010visual}. 
Overall, the proposed algorithm performs favorably against the other trackers, e.g., DGT (Our previous version) (PR/SR: 4.8\%/3.2\%), Staple (8.4\%/3.7\%) and SRDCF (10.0\%/5.4\%). 
%

\subsection{Evaluation on the NUS-PRO Dataset}
We evaluate the proposed algorithm against the state-of-the-art trackers on the NUS-PRO~\cite{NUS-PRO16pami} dataset.

\subsubsection{Overall Performance}
We present the evaluation results of our method against 20 conventional trackers
on the NUS-PRO dataset ~\cite{NUS-PRO16pami} in Figure~\ref{fig::NUS-PRO_plots}. 
Overall, the proposed tracker performs favorably against other trackers on the NUS-PRO dataset. 
The results of the top 4 performing methods (CPF~\cite{Perez02eccv}, ASLA~\cite{Jia12cvpr}, SCM~\cite{SCM12cvpr} and LOT~\cite{Oron12cvpr}) show that the combination of 
local feature representations and particle filter search models can achieve the state-of-the-art 
performance. 
Although adopting only the local feature representation, the proposed tracking algorithm performs well on the NUS-PRO dataset.

\subsubsection{Category-based Evaluation}
We present how the proposed tracker performs on 4 object categories in the NUS-PRO database.
The AUC plots of TRR curves in Figure~\ref{fig::NUS-PRO_category}
show that 
the proposed method performs well in the \emph{rigid object}, \emph{sportsman} and \emph{face} sequences, and comparably with the SCM scheme 
in the \emph{pedestrian} sequences. 
The sportsman category is the most challenging among 4 object types in the NUS-PRO database, followed by the classes of \emph{pedestrians}, \emph{rigid objects} and \emph{faces}. 
%

\begin{table}[t]
\centering
\caption{Comparison of WPG against the SOWP method~\cite{Kim15iccv} and the top three trackers on the VOT2014 challenge dataset~\cite{vot14challenge}. `ACC w/o' denotes the ACC score without the re-initialization step. The best performance is in boldface. }
\label{tb::vot14}
\centering
\begin{tabular}{c | c c c | c c c}
\hline
\hline
 & \multicolumn{3}{c|}{Baseline} & \multicolumn{3}{c}{Region noise}\\
\hline
 & ACC  & ROB & ACC w/o & ACC  & ROB & ACC w/o \\
\hline
DSST & {\bf 0.66}  & 1.16 & 0.46 & {\bf 0.57} & 1.28 & 0.43 \\
SAMF & 0.61  & 1.28 & 0.50 & {\bf 0.57} & 1.43 & 0.43 \\
KCF & 0.62  & 1.32 & 0.39 & {\bf 0.57} & 1.51 & 0.36 \\
SOWP & 0.57  & 0.56 & 0.51 & 0.55 & 0.68 & 0.48 \\
\hline
WPG & 0.57  & {\bf 0.53} & {\bf 0.52} & 0.55 & {\bf 0.50} & {\bf 0.50} \\
\hline
\hline
\end{tabular}
\end{table}

\subsubsection{Attribute-based Evaluation}
We present the AUC plots of TRR curves of the evaluated tracking algorithms based on 12 attributes, including shadow change (SC), flash (FL), dim light (DL), camera shaking (CS), scale change (SC), rotation (RO), shape deformation (SD), partial occlusion (PO), full occlusion (FO), clutter background (CB), similar objects (SO) and fast background change (FBC). 
The proposed tracker performs well on scenes with most attributes including SC, DL, CS, SC, RO, SD, PO, FO, CB and SO. 
The evaluation results are consistent with the findings on the OTB100 dataset except that 
the FL, DL, CS and FBC attributes are not reported on the OTB100 dataset
and the proposed method performs slightly worse than the others on the scenes with the DL and FBC attributes. 
The performance on the sequences with the DL attribute may be explained by the adopted
features (color and gradient) of the proposed method for representing target objects 
under low illumination conditions, which can be improved 
by using integrating more features~\cite{Li16tip1}.
On the other hand, the performance of the proposed algorithm on sequences with  the FBC 
attribute can be explained by the search strategy, and can be further improved by using robust motion or search models to leverage more temporal and spatial information.

{\subsection{Evaluation on the VOT Challenge Dataset}
Finally, we report the evaluation results of WPG against SOWP~\cite{Kim15iccv} and the top three trackers (i.e., DSST~\cite{ScaleCSK14bmvc}, SAMF~\cite{Agapito14eccv} and KCF~\cite{CSK15pami}) on the VOT2014 challenge dataset~\cite{vot14challenge}, as shown in Table~\ref{tb::vot14}. In Baseline evaluation, a tracker is initialized with a ground truth. In Region noise evaluation, a tracker inputs a perturbed ground truth.}

{From Table~\ref{tb::vot14}, we can see that WPG obtains low ACC scores, achieves the best ROB results in both evaluations. In the VOT challenge, a re-initialization step is triggered using a new ground truth when a tracker is detected as failure. Therefore, the compared trackers fail to track more frequently than WPG, and thus they obtain higher overlap ratios. To mitigate these effects of re-initialization, we remove re-initialization step in evaluations, and denote overlap ratios as ACC w/o. The results show that WPG yields the best ACC scores without the re-initialization.}

{It is worth noting that sequences of Region noise evaluation contain more clutter and noise, but WPG performs better on Region noise evaluation than on Baseline. It suggests that WPG can handle region noise more effectively than others.}

\begin{table}[t]
\centering
\caption{Performance of 4 variants of the proposed method against the SOWP method~\cite{Kim15iccv}. }
\label{tb::component}
\centering
\begin{tabular}{c|c|c c c c}
\hline
\hline
 & SOWP & WPG$_A$ & WPG$_Z$ & WPG$_E$ & WPG$_W$ \\
\hline
PR & 0.803 & 0.870 & 0.882 & 0.873  & 0.811 \\

SR & 0.560 & 0.610 & 0.626 & 0.612  & 0.597 \\
\hline
\hline
 & WPG & WPG$_A'$ & WPG$_Z'$ & WPG$_E'$ \\
\hline
PR & 0.894   & 0.861   & 0.883 & 0.878\\

SR & 0.632   & 0.612   & 0.627 & 0.624\\
\hline
\hline

\end{tabular}
\end{table}

\subsection{Analysis}
To demonstrate the effectiveness of the main components, we present empirical results using 4 variants of the proposed algorithm on the OTB100 dataset . 
%
These variants are: 1) WPG$_W$:  We remove the patch weights in our tracking algorithm, 
2) WPG$_A$: We remove the affinity learning and directly utilize the representation coefficients to diffuse patch weights.
The objective function is:
\begin{equation}
\label{eq:graph_model3}
\begin{aligned}
&\underset{{\bf Z}, {\bf E}, {\bf w}}{\min}~\alpha||{\bf Z}||_0+\beta||{\bf E}||_{2,0}+\lambda_1\sum_{i,j}{{\bf z}_{ij}({\bf w}_i-{\bf w}_j)^2}\\
&+\frac{\lambda_2}{2}||\Gamma\circ({\bf w}-{\bf r})||^2+\frac{\lambda_4}{2}||{\bf w}||^2\\
&\mbox{s.t.} \quad {\bf X}={\bf XZ} + {\bf E},~{\bf w}\geq 0.
\end{aligned}
\end{equation}
We also use the ADMM algorithm to solve~\eqref{eq:graph_model3}. 
%
3) WPG$_Z$: We remove the sparse constraints on ${\bf Z}$, but enforce minimizing $||{\bf Z}||_F^2$ to 
avoid the trivial solution. 
Thus, ${\bf Z}$ can be updated with the closed-form solution:
\begin{equation}
\label{eq:updateZ1}
\begin{aligned}
&{\bf Z}^{k+1}=(\frac{\alpha+\mu^k}{\mu^k}{\bf I}+{\bf X}^{\top}{\bf X})^{-1},\\
&({\bf X}^{\top}({\bf X}-{\bf E}^k+\frac{{\bf Y}_1^k}{\mu^k})+{\bf Q}^k-\frac{{\bf Y}_2^k}{\mu^k}).
\end{aligned}
\end{equation}
4) WPG$_E$: We remove the sample-specific sparse constraints on ${\bf E}$ but  
enforce minimizing $||{\bf E}||_F^2$ to avoid the trivial solution. 
Thus, ${\bf E}$ can be updated with a closed-form solution:
\begin{equation}
\label{eq:updateE1}
\begin{aligned}
{\bf E}^{k+1}=\frac{\mu^k}{\mu^k+\beta}({\bf X}-{\bf XZ}^{k+1}+\frac{{\bf Y}_1^k}{\mu^k}).
\end{aligned}
\end{equation}

{To rules out the implementation flaw or optimization differences, we set parameters $\gamma$, $\alpha$ and $\beta$ to a ridiculously low number (e.g., $10^{-10}$ in this work) to render contribution of each term, and denote them as WPG$_A'$, WPG$_Z'$ and WPG$_E'$, respectively.}

Table~\ref{tb::component} shows the evaluation results against the SOWP method~\cite{Kim15iccv}. 
The performance gains achieved by the proposed algorithm over the SOWP method demonstrate 
the significances of the main components. 
In particular, the results show that:
1) Introducing patch weights into the object representations helps suppress the effects of background clutters 
in visual tracking by comparing the performance of WPG$_W$ against the other schemes. {WPG is a spatially reliability learning method, which has been proven to be an effective way to mitigate outlier effects, and thus bring big performance gains for achieving state-of-the-art tracking performance~\cite{dcf-csr16cvpr,Sun18cvpr}.}
2) The WPG, WPG$_A$, WPG$_Z$ and WPG$_E$ methods perform well against the SOWP scheme,  
which suggests that the dynamic graph facilitates optimizing the patch weights by capturing the intrinsic relationship among image patches. {Comparing with restriction of spatial neighbors in SOWP, variants of WPG are good at exploring long-range relationships among patches, and also mitigating noise effects of low-level features. Hence, the patch weights optimized by WPG variants are more accurate and robust.}
3) The WPG algorithm performs better than the WPG$_A$, WPG$_Z$ and WPG$_E$ schemes, thereby 
justifying the effectiveness of learning graph affinity matrix ${\bf A}$,  
sparse constraints on ${\bf Z}$, and 
sample-specific sparse constraints on ${\bf E}$, respectively. 
{First, sparse representation based graph~\cite{Wright09pami,Wright10procIEEE} could automatically select the most informative neighbors for each patch, and explore higher order relationships among patches, hence is more powerful and discriminative. Second, the learned graph could suppress corrupted and noisy image patches by modelling noise in sparse representation.}

{The performances of WPG$_A'$, WPG$_Z'$ and WPG$_E'$ against WPG further demonstrate above observations and conclusions.}

\section{Conclusions}
\label{sec::conclusion}
In this paper, we propose an effective algorithm for visual tracking by suppressing the effects of background clutters.
A patch-based graph is learned dynamically by capturing the intrinsic relationship among patches. 
To reduce the computational complexity, we develop an efficient algorithm for the proposed model by solving several convex subproblems. 
Finally, the optimized patch weights are incorporated into the structured SVM framework to carry out 
the tracking task.
Extensive experimental results on three benchmark datasets demonstrate the effectiveness of the proposed algorithm  over the state-of-the-art methods. 
Our future work will focus on: 1) learning the dynamic spatio-temporal graphs to explore more relations among image patches, 
2) developing robust motion or search models for addressing fast object or background motions, 
and 3) replacing the hand-craft features with hierarchical appearance models for more effective representations.





\ifCLASSOPTIONcaptionsoff
  \newpage
\fi

\bibliographystyle{IEEEtran}
\bibliography{tpami17}

\begin{IEEEbiography}[{\includegraphics[width=1in,height=1.25in,clip,keepaspectratio]{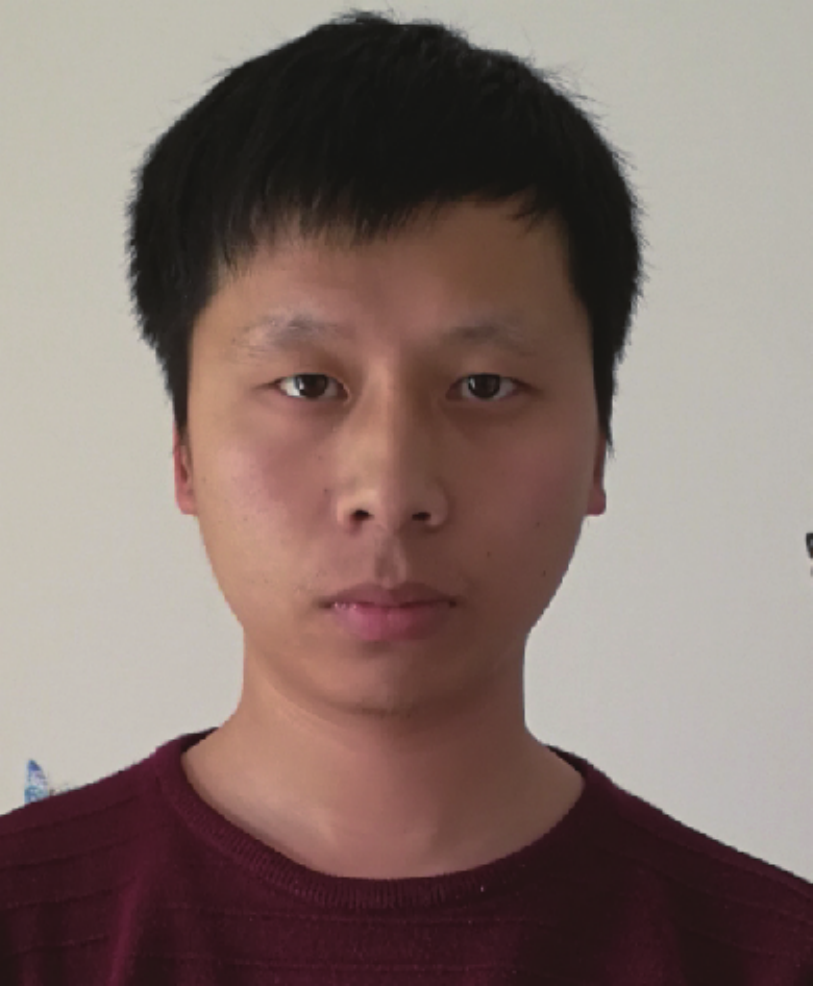}}]{Chenglong Li} 
received the M.S. and Ph.D. degrees
from the School of Computer Science and Technology,
Anhui University, Hefei, China, in 2013 and
2016, respectively. From 2014 to 2015, he was a visiting
student with the School of Data and Computer
Science, Sun Yat-sen University, Guangzhou, China.
He is currently a lecturer at the School of Computer
Science and Technology, Anhui University, and also
a postdoctoral research fellow at the Center for Research on Intelligent Perception and Computing (CRIPAC), National Laboratory
of Pattern Recognition (NLPR), Institute of
Automation, Chinese Academy of Sciences (CASIA),
China. He was a recipient of the ACM Hefei Doctoral Dissertation Award in
2016.

\end{IEEEbiography}

 \begin{IEEEbiography}[{\includegraphics[width=1in,height=1.25in,clip,keepaspectratio]{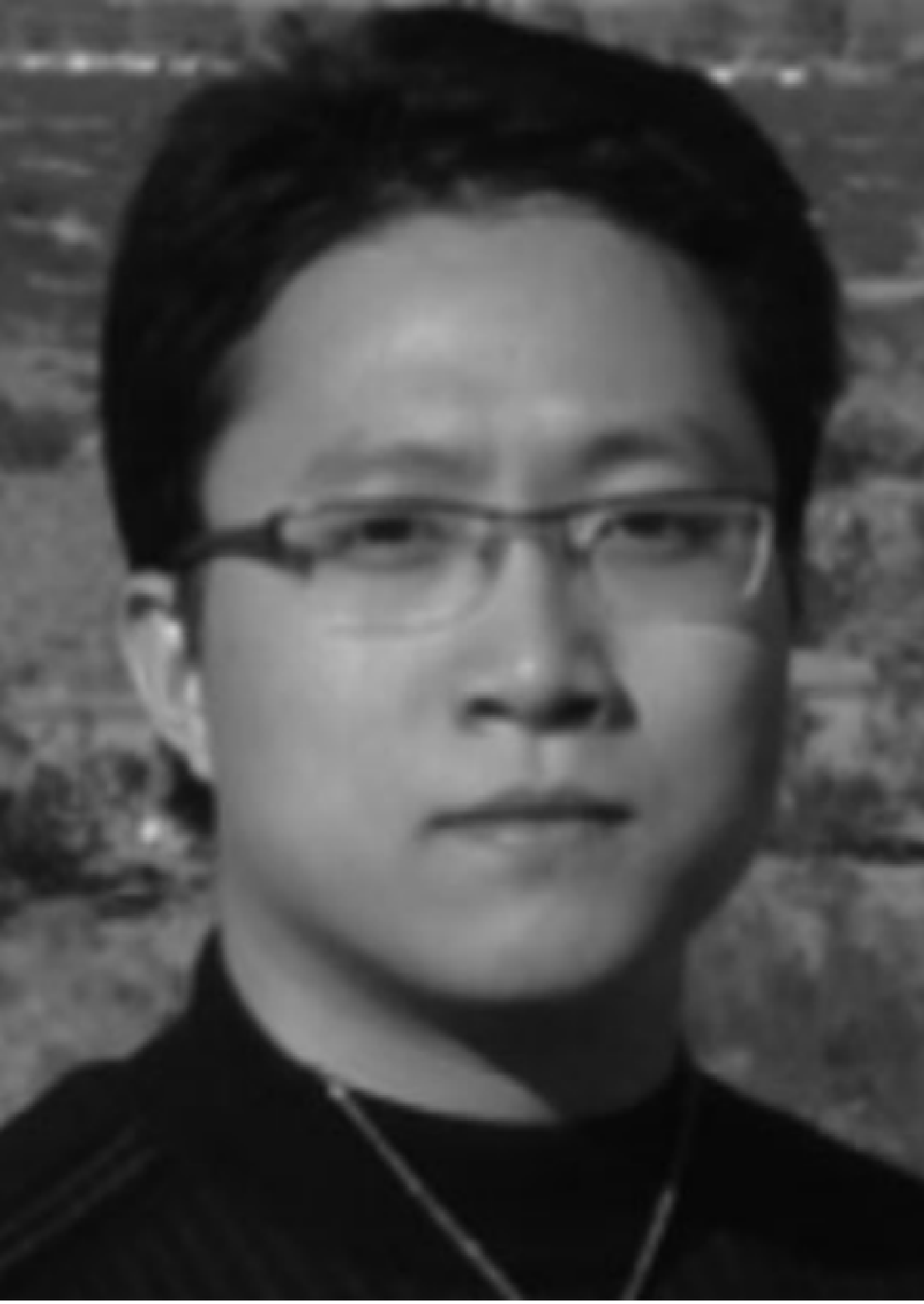}}]{Liang Lin}
received the BS and PhD degrees
from the Beijing Institute of Technology (BIT),
Beijing, China, in 1999 and 2008, respectively.
He is currently a full Professor with the School of
Advanced Computing, Sun Yat-Sen University,
Shunde, China.
From 2006 to 2007, he was a joint PhD student with the Department of Statistics, University of California, Los Angeles (UCLA), CA. He was a Post-Doctoral Research Fellow with the Center for Vision, Cognition, Learning, and Art of UCLA.
He was
supported by several promotive programs or funds for his work such
as Program for New Century Excellent Talents of Ministry of Education
(China) in 2012 and Guangdong Distinguished Young Scholar Fund in
2013. Dr. Lin was a recipient of Best Paper Runners-Up Award in ACM
NPAR 2010, Google Faculty Award in 2012, and Best Student Paper
Award in IEEE ICME 2014.
\end{IEEEbiography}

\begin{IEEEbiography}[{\includegraphics[width=1in,height=1.25in,clip,keepaspectratio]{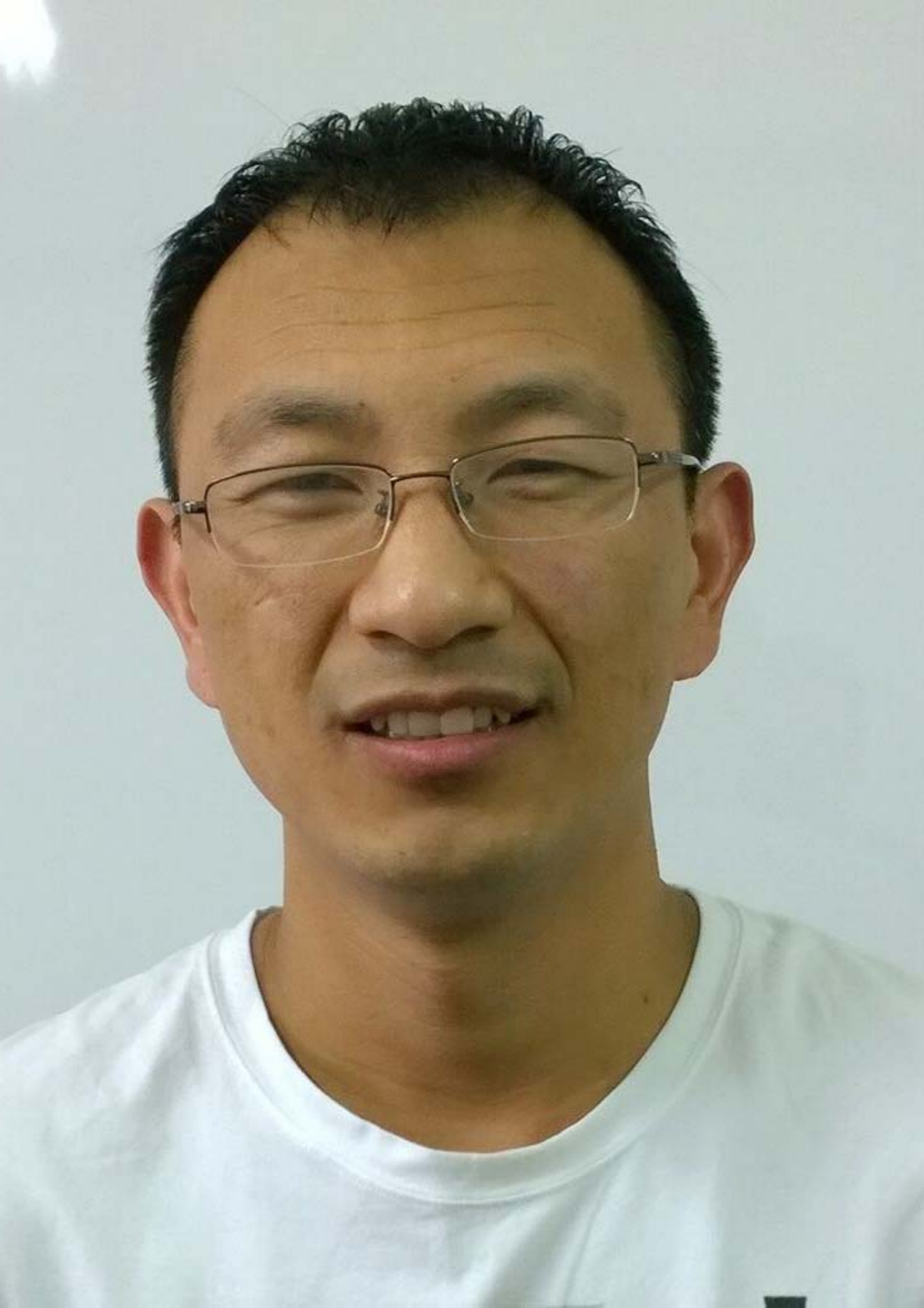}}]{Wangmeng Zuo}
received the Ph.D. degree in computer application technology from the Harbin Institute of Technology, Harbin, China, in 2007.
From August 2009 to February 2010, he was a Visiting Professor in Microsoft Research Asia.
He is currently a Professor in the School of Computer Science and Technology, Harbin Institute of Technology. His current research interests include image modeling and blind restoration, discriminative learning, biometrics, and computer vision.
Dr. Zuo is an Associate Editor of the IET Biometrics.
\end{IEEEbiography}

\begin{IEEEbiography}[{\includegraphics[width=1in,height=1.25in,clip,keepaspectratio]{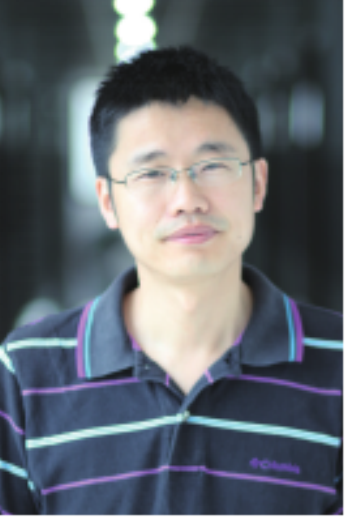}}]{Jin Tang} 
received the B.Eng. degree in the School of Automation
and the Ph.D. degree in the School of Computer Science and Technology at
Anhui University, Hefei, China, in 1999 and 2007,
respectively.
He is currently a Professor in the School of Computer Science and Technology at Anhui University. 
\end{IEEEbiography}

\begin{IEEEbiography}[{\includegraphics[width=1in,height=1.25in,clip,keepaspectratio]{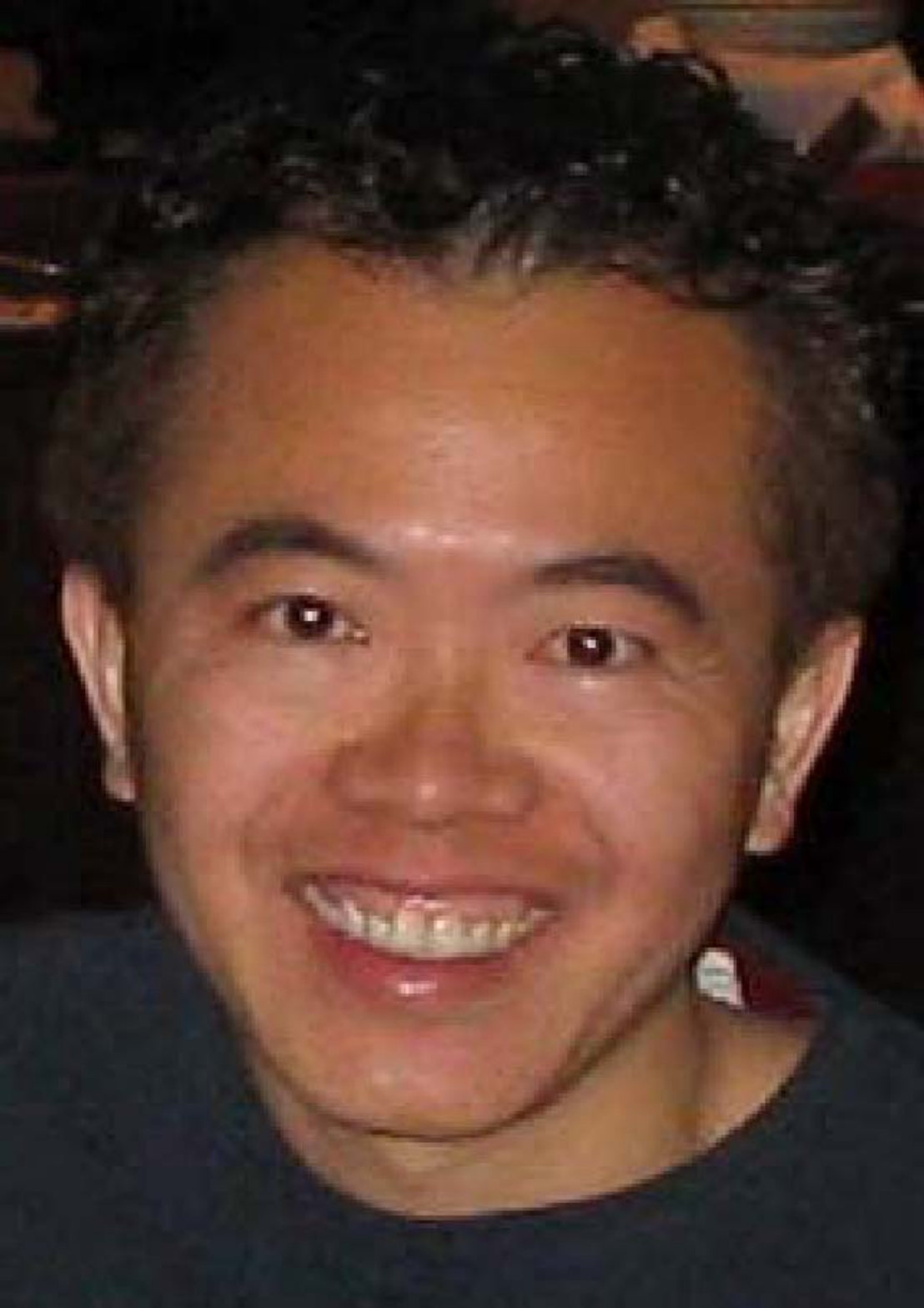}}]{Ming-Hsuan
    Yang} is a professor in Electrical Engineering and
  Computer Science at University of California, Merced. He received
  the PhD degree in computer science from the University of Illinois
  at Urbana-Champaign in 2000.
Yang served as an   associate editor of the IEEE Transactions on Pattern Analysis and
  Machine Intelligence from 2007 to 2011, and is an associate editor
  of the International Journal of Computer Vision, Image and Vision
  Computing and Journal of Artificial Intelligence Research. He
  received the NSF CAREER award in 2012
and the Google Faculty Award in 2009.
He is a senior member of the IEEE and the ACM.
\end{IEEEbiography}

\end{document}